\documentclass[preprint,12pt]{elsarticle}

\usepackage[colorlinks=true,linkcolor=blue,urlcolor=blue,citecolor=blue]{hyperref}
\usepackage{amssymb}
\usepackage{amsmath}
\usepackage{cleveref}
\usepackage{color}
\usepackage{url}
\usepackage{tcolorbox}
\usepackage{tabularx}
\usepackage{colortbl}

\definecolor{green}{HTML}{44AA99}
\definecolor{yellow}{HTML}{DDCC77}
\definecolor{blue}{HTML}{88CCEE}
\definecolor{red}{HTML}{CC6677}
\definecolor{darkred}{HTML}{DC3220}

\usepackage{array}
\usepackage{listings}
\usepackage[T1]{fontenc}
\usepackage{multirow}
\usepackage{mwe}
\usepackage[export]{adjustbox}
\usepackage{tabularray}

\begin{document}

\makeatletter
\def\ps@pprintTitle{%
  \let\@oddhead\@empty
  \let\@evenhead\@empty
  \def\@oddfoot{\reset@font\hfil\thepage\hfil}
  \let\@evenfoot\@oddfoot
}
\makeatother

\begin{frontmatter}

\title{Exploring Vision Language Models for Facial Attribute Recognition: Emotion, Race, Gender, and Age}

\author[label1]{Nouar AlDahoul}\ead{naa9497@nyu.edu}
\author[label2]{Myles Joshua Toledo Tan}\ead{tan.m@ufl.edu}
\author[label2]{Harishwar Reddy Kasireddy}\ead{harishwarreddy.k@ufl.edu}
\author[label1]{Yasir Zaki\texorpdfstring{\corref{cor1}}{}}\ead{yasir.zaki@nyu.edu}

\cortext[cor1]{Corresponding author.}

\affiliation[label1]{organization=Computer Science Department, New York University Abu Dhabi,city=Abu Dhabi,country=UAE}

\affiliation[label2]{organization=Department of Electrical and Computer Engineering, Herbert Wertheim College of Engineering, University of Florida,
            city=Florida,
            country=USA}
            


\begin{abstract}
 Technologies for recognizing facial attributes like race, gender, age, and emotion have several applications, such as enhanced surveillance, personalized advertising content, sentiment analysis, interactive media, and the study of demographic trends and social behaviors. Analyzing demographic characteristics based on images and analyzing facial expressions have several challenges due to the complexity of humans' facial attributes and diversity in representation. Traditional approaches have employed convolutional neural networks (CNNs) and various other deep learning techniques, trained on extensive collections of labeled images. While these methods have demonstrated effective performance, there remains potential for further enhancements to increase the recognition accuracy. In this paper, we propose to utilize vision language models (VLMs) such as generative pre-trained transformer (GPT), Google GEMINI, large language and vision assistant (LLAVA), Google PaliGemma (Pathways Language and Image Model + Gemma model), and Microsoft Florence2 to recognize facial attributes such as race, gender, age group, and emotion from images with human faces. Various datasets like FairFace, AffectNet, and UTKFace have been utilized to evaluate the proposed solutions. The results show that VLMs are competitive and sometimes superior to traditional techniques. Additionally, we propose ``FaceScanPaliGemma''---a fine-tuned PaliGemma model---for race, gender, age, and emotion recognition tasks. The results show an accuracy of 81.1\% , 95.8\%, 80\%, and 59.4\% for race, gender, age group, and emotion classification, respectively, outperforming pre-trained version of PaliGemma, other VLMs, and state-of-the-art methods. Finally, we propose ``FaceScanGPT'', which is a GPT-4o model to recognize race, gender, age group, and emotion when several individuals are present in the image using a prompt engineered for a person with specific facial and/or physical attributes. The results underscore the superior multitasking capability of FaceScanGPT to detect the individual's physical attributes like hair cut, clothing color, actions performed, postures, etc., and facial attributes like race, gender, etc., using only a prompt to drive the detection and recognition tasks. 

\end{abstract}





\end{frontmatter}

\section{Introduction}

Technologies for recognizing attributes such as race, gender, age, and emotion have various applications, including improved surveillance and personalized advertising content. Face attribute recognition is also utilized in demographic surveys conducted in marketing or social science research, with the goal of understanding human social behaviors and their connections to individuals' demographic backgrounds~\cite{karkkainen2021fairface}. Accurately identifying multiple attributes in an image presents several challenges due to the complexity of human attributes and diversity in representation. Human attributes like race, gender, age, and emotion are not always well-defined and can be subjective~\cite{yuan2024multi}. Machine learning (ML) models heavily rely on training data, and often biased or unrepresentative datasets can lead to biased or inaccurate predictions~\cite{vaidya2020empirical}. Attributes like age and emotion exhibit significant intra-class variability, meaning that different individuals within the same category (e.g., the same age group or emotion class) can have vastly different visual appearances~\cite{lee2014intra}. 

In the area of image processing, artificial intelligence (AI), particularly ML, has made remarkable strides, transforming how computers understand and analyze visual data. Multi-task learning allows a single model to tackle multiple related tasks simultaneously, leveraging shared knowledge to improve performance across the board~\cite{caruana1997multitask}. In image processing, multi-task learning becomes indispensable, especially in the space of multi-attribute classification. Generative models represent a revolutionary step forward in artificial intelligence, particularly in the areas of natural language processing, vision, and generation. These models are designed to understand and generate human-like content, ushering in a new era of AI capabilities and applications.

Maintaining separate models for each individual task results in increased computational overhead and model complexity. Single-task classifiers are optimized to excel at a specific task but may struggle to generalize in unseen data or adapt to new tasks without retraining~\cite{li2006improving}. Many tasks in image processing share common underlying patterns and features. Single-task classifiers fail to leverage this shared information effectively, leading to suboptimal performance and redundant learning~\cite{han2017heterogeneous}. For instance, both facial recognition and facial attribute analysis tasks benefit from learning low-level visual features like edges and textures. 

The critical role of ethics in AI development cannot be overstated, especially when it comes to applications involving sensitive human attributes such as race, gender, age, and emotion~\cite{mohammad2022ethics,sham2023ethical}. As AI technologies become increasingly integrated into various aspects of society, it is imperative to prioritize ethical considerations to ensure fairness, transparency, and accountability. Many datasets used today to train AI models suffer from various biases, reflecting historical inequalities and societal prejudices~\cite{leslie2020understanding,li2020deeper}. Even when trained on unbiased datasets, AI algorithms can perpetuate or amplify existing biases~\cite{schmitz2022bias}. This can occur due to factors such as feature selection, model architecture, or optimization objectives. 

The aim behind this work is to develop a multi-task classifier that leverages the capabilities of vision language models such as GPT~\cite{brown2020language}, GEMINI~\cite{team2023gemini}, LLaVA~\cite{liu2024visual}, PaliGemma~\cite{beyer2024paligemma}, and Florence-2~\cite{xiao2024florence} generative models to simultaneously identify multiple human facial attributes---specifically race, gender, age, and emotion---from given images. This classifier integrates the advanced generative and understanding capabilities of these models to accurately and efficiently process and classify complex human's facial characteristics across diverse datasets. 
The creation of this multi-task classifier takes into account incorporating ethical AI practices, concentrating on reducing biases that new vision language models have recently addressed, ensuring privacy and consent, maintaining transparency and accountability, and promoting diversity and inclusion.

The use of VLM generative models significantly enhances the learning capabilities of the proposed multi-task classifier. These models bring several key advantages:
\begin{enumerate}
    \item VLMs such as GPT ~\cite{brown2020language}, GEMINI~\cite{team2023gemini}, LLaVA~\cite{liu2024visual}, PaliGemma~\cite{beyer2024paligemma}, and Florence-2~\cite{xiao2024florence} are designed to capture deep, nuanced representations of data. VLM, with its sophisticated language understanding capabilities, excels at interpreting and generating both textual and image-based information~\cite{kevian2024capabilities}. 
    \item The pre-trained nature of these models utilizes transfer learning approach to offer a robust foundation of learned features that can be fine-tuned to specific tasks, such as the simultaneous classification of race, gender, age, and emotion~\cite{dalvi2021survey}.
    \item VLM's architecture is particularly adept at understanding context, a critical factor when dealing with attributes that may be subtly expressed or highly context-dependent~\cite{demszky2023using}.
    \item The models are scalable and flexible, allowing for continuous updates and adaptations as new data becomes available or as requirements evolve. This adaptability is crucial for keeping the classifier relevant and effective across different populations and evolving societal norms~\cite{casado2023ensemble}.
\end{enumerate}

The approach of using vision language models for the multi-task classification of human attributes like race, gender, age, and emotion offers several significant benefits that address both the efficiency and effectiveness of AI systems:
\begin{enumerate}
    \item Employing multi-task learning with vision language models leads to classifying multiple attributes simultaneously, rather than requiring separate models for each attribute, which helps to reduce computational resources~\cite{minaee2024large}.
    \item VLM's advanced capabilities in handling multi-modal data and their profound contextual understanding enable them to adapt to the real-world diversity found in human attributes~\cite{li2024multimodal}.
    \item In traditional setups where separate models are trained for each task, there is often a significant overlap in what each model learns. VLM plays a crucial role in reducing this redundant learning~\cite{brown2020language}.
\end{enumerate}

In this study, our proposed facial attribute recognition solutions utilize state-of-the-art vision language models to recognize race, gender, age group, and emotion of persons using their face images. Our contributions can be summarized as follows:

\begin{itemize}
    \item We formulated a facial attribute recognition task as a visual question answering task using various VLMs.
    \item We utilized public image datasets such as FairFace, AffectNet, and UTKFace, that have face images for various races, genders, age groups, and emotions for evaluation and comparison.
    \item We explored and evaluated the zero-shot classification, which is a significant capability existing in vision language models such as Google GEMINI 1.5, GPT-4o, LLAVA-NEXT, PaliGemma, and Florence-2, and employed it in the task of human's facial attribute recognition.
    \item We fine-tuned PaliGemma utilizing FairFace and AffectNet datasets to improve the recognition accuracy. The outcome is FaceScanPaliGemma VLM.
    \item We explored the multitasking capability of GPT-4o using images with several persons having various physical and facial attributes.
\end{itemize}

This rest of the paper is organized as follows: In Section~\ref{sec:relatedwork}, we review previous works on facial attribute recognition methods. Section~\ref{sec:motivation} presents our research motivation. In Section~\ref{sec:methods}, we describe the datasets used to run the experiments. Section~\ref{sec:results} discusses the experimental results and compares the proposed solution with other baseline methods. Finally, conclusions and future works are discussed in Section~\ref{sec:conclusion}.

\section{Related Work}
\label{sec:relatedwork}

\subsection{Challenges in facial attribute datasets}

The advancements in gender, race, age, and emotion classification applications requires datasets of diverse facial images that can address challenges such as imbalanced samples, pose variations, and varying lighting conditions. Recently, CNNs, especially the FaceNet model~\cite{schroff2015facenet}, have shown robustness in handling unbalanced data distributions~\cite{mustapha2021age}.
In previous works, they identified a significant bias in public face image datasets, which predominantly feature Caucasian faces while considerably under-representing other racial groups, such as Latinx~\cite{karkkainen2021fairface}. To address this issue of racial imbalance, they compiled a dataset, namely FairFace, consisting of 108,501 face images that is balanced across different races~\cite{karkkainen2021fairface}. They used this dataset to train a ResNet-34~\cite{he2016deep} model to evaluate the classification performance for gender, race, and age. Similarly, recent works~\cite{aldahoul2024ai} used the same dataset with different models to improve the accuracy of the classification. They utilized VGGFace ResNet-50 convolutional neural network (VGGFace ResNet-50 CNN)~\cite{cao2018vggface2} to extract the embedding vector from the face images in the FairFace dataset~\cite{aldahoul2024ai}. They added support vector machine (SVM) classifier after removing the top layers. Additionally, they explored numerous models including FaceNet+SVM~\cite{schroff2015facenet}, tuned EfficientNet-B7~\cite{DBLP:journals/corr/abs-1905-11946}, and large vision transformer~\cite{dosovitskiy2020image} for race, and gender classification~\cite{aldahoul2024ai}. Furthermore, CLIP's zero-shot classifier  demonstrated their performance for race and gender classification using the FairFace dataset~\cite{radford2021learning}. 

Existing annotated databases of facial expressions in the wild are small and include clean and high-quality posed facial expressions. However, posed expressions may not accurately represent many of the unopposed facial expressions encountered in daily life. As a result, a dataset, namely AffectNet~\cite{mollahosseini2008affectnet}, was proposed to be the largest database of facial expressions with annotations for eight expressions. Several works in literature have utilized the AffectNet dataset to train and evaluate their models for emotion or facial expression recognition~\cite{ning2024representation,mao2023poster++,savchenko2022classifying,li2022emotion}.

\subsection{Deep Learning for facial attributes recognition}

Race classification has become significant in applications like surveillance~\cite{abdulwahid2023classification} and market advertising~\cite{advertising}. Recently, various deep learning models have been used for race identification~\cite{abdulwahid2023classification,sunitha2022intelligent,ahmed2022race,al2021classification,albdairi2022face}. 

Similarly, for gender recognition, deep learning techniques have improved the gender classification accuracy~\cite{haseena2022prediction,fayyaz2023pedestrian,sonthi2023deep,yaman2022multimodal,tunc2022age,ciobotaru2023comparing}. Based on these techniques, the effectiveness of CNNs and AlexNet~\cite{krizhevsky2012imagenet} under challenging conditions has been demonstrated, showing potential for applications of tracking and identification~\cite{sonthi2023deep}. 

Additionally, age estimation from facial images is crucial for applications like security and social interaction. Several deep learning techniques have been proposed in the literature to improve feature extraction and thus enhance the accuracy of age classification~\cite{haseena2022prediction,yaman2022multimodal,reddy2020age,duan2018hybrid}. In this context, hybrid deep learning structures that combine CNNs and Extreme Learning Machine~\cite{huang2006extreme} have confirmed the efficacy of integrating multiple learning strategies~\cite{duan2018hybrid}.

Emotion recognition from facial expressions is crucial in human-computer interaction, with deep learning methods addressing challenges such as pose variations, illumination changes, and occlusions~\cite{chowdary2023deep}. Based on this, transfer learning techniques using networks like ResNet50~\cite{he2016deep} and VGG19~\cite{simonyan2014very} have proven effective in combining feature extraction and classification for emotion recognition task~\cite{chowdary2023deep}. 

Multimodal or multitask deep models have shown improvement in classification accuracy, demonstrating the utility of leveraging diverse biometric inputs~\cite{yaman2022multimodal,tunc2022age,ciobotaru2023comparing}. The work aim to integrate multiple facial attributes extracted from facial images. This integration  is crucial and has become a key area of interest in computer vision applications. CNN has been utilized to integrate gender and emotion~\cite{pandi2022emotion}. Similarly, Visual attention-driven architectures have been employed for gender and ethnicity integration~\cite{khellat2022gender}. Additionally, the integration of age and gender prediction has enhanced biometric security and personalized systems. 

Several studies have highlighted the efficiency of deep learning in enhancing real-time biometric recognition, paving the way for more compact, faster, and accurate systems. A multi-task CNN has been proposed for recognizing gender, age, ethnicity and emotion~\cite{foggia2023multi}. It improved efficiency in processing and memory usage while maintaining good accuracy across multiple tasks making it suitable for embedded systems~\cite{foggia2023multi} and for applications requiring real-time processing with limited computational resources~\cite{pandi2022emotion}. 

Previously mentioned deep learning methods have outperformed traditional techniques in classifying race, gender, age, and emotion. However, there is still room for improvement. As such, this paper aims to improve facial attribute classification using VLMs.

\subsection{Emerging role of LLMs in image processing}

While LLMs are widely known for their expertise in language processing, recent studies have begun to explore their effectiveness in image recognition, often through multi-modal learning approaches~\cite{abdelhamed2024you}. Multimodal models like CLIP (contrastive language-image pretraining)~\cite{radford2021learning} leverage the combination of LLMs and CNNs by learning from pairs of images and their textual descriptions. CLIP has achieved good performance on various image recognition benchmarks, including race, gender, and age recognition~\cite{radford2021learning}, thanks to its ability to create a shared representation space for both images and text, which enables robust performance even in zero-shot settings.

The use of large language models (LLMs), such as GPT~\cite{Hello_GPT-4o} and BERT~\cite{devlin2018bert}, has extended beyond text-based tasks, delving into areas like image recognition and processing. This broadening of application stems from the models' capacity to comprehend and generate human-like text, offering a fresh perspective on how images can be interpreted and analyzed~\cite{radford2021learning}.

Recent interdisciplinary research has started to investigate the potential of LLMs in image-related tasks such as construction processes, radiology, and medical visual question answering~\cite{yang2024vision,yildirim2024multimodal,hartsock2024vision}. For instance, studies have shown that GPT can generate textual descriptions from images, paving the way for innovative approaches to image understanding through natural language~\cite{radford2021learning}.

LLMs have also been integrated with traditional vision models in tasks like visual question answering (VQA)~\cite{antol2015vqa}, where models are trained to respond to questions based on image content. This highlights the synergy between LLMs and image recognition, requiring a deep understanding of both visual and textual data~\cite{antol2015vqa}.

Integrating Large Language Models (LLMs) such as Google GEMINI 1.5~\cite{Gemini_15_technical_report,Introducing_Gemini_15}, GPT-4o~\cite{Hello_GPT-4o,GPT-4o}, LLAVA-NEXT~\cite{LLaVA_paper,LLaVA-NeXT_paper,Visual_and_Language}, PaliGemma~\cite{beyer2024paligemma}, and Florence-2~\cite{xiao2024florence} into facial recognition tasks can significantly enhance the performance and capabilities of deep learning models. This paper investigates the use of vision language models (VLMs) in addressing challenging computer vision tasks, such as emotion, gender, race, and age group classification. It leverages VLMs' capabilities in understanding and generation to improve visual recognition.

\section {Research Motivation}
\label{sec:motivation}
The direct applications of vision language models (VLMs) in facial attribute recognition are still unexplored. The adaptability and contextual comprehension of VLMs hold promise for tackling complex challenges in facial attribute recognition, such as identifying race, gender, age, and emotions in face images that are blurred, noisy, have varying illumination, or have different facial orientations.

The existing solutions of VLMs~\cite{yang2024vision,yildirim2024multimodal,hartsock2024vision,zhu2024harnessing} in various applications indicate, with their sophisticated understanding and generative abilities, their potential capability to serve as complementary---or even alternative---solutions to traditional and CNN-based methods in facial attribute recognition systems. By harnessing the advanced language processing and contextual analysis capabilities of VLMs, researchers have the potential to achieve significant improvements in the accuracy, efficiency, and adaptability of technologies used for recognizing race, gender, age, and emotion.

\section{Materials and Methods}
\label{sec:methods}
This section describes the datasets used in the experiments conducted to evaluate the VLMs performance on facial attribute recognition tasks. Moreover, the section discusses the baseline methods usually used in the literature and highlights our proposed solutions. 

\subsection{Dataset Overview}

\subsubsection{FairFace dataset}
\begin{figure}[!htb]
    \centering
    \includegraphics[width=1\linewidth]{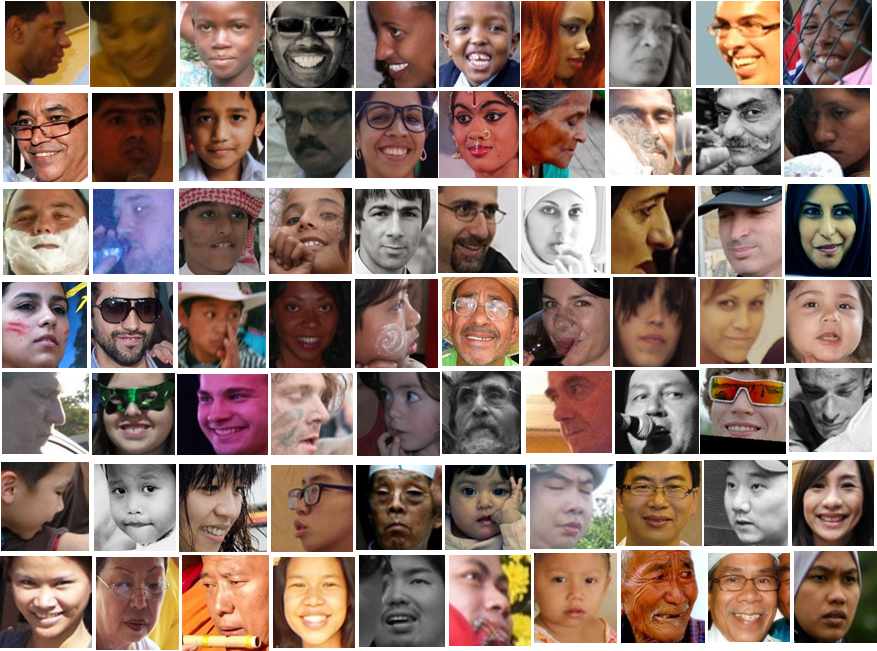}
    \caption{Several samples from each race category to show the challenging contents of this dataset. The races are as follows: first row: Black, second row: Indian, third row: Middle Eastern, fourth row: Latinx\_Hispanic, fifth row: White, sixth row: East Asian, and seventh row: Southeast Asian.
    }
    \label{fig:FairFace}
\end{figure}

To address the issue of racial bias in existing datasets, a face image dataset was created comprising of 108,501 images that is racially balanced~\cite{karkkainen2021fairface}. The images were categorized into seven racial groups: White, Black, Indian, East Asian, Southeast Asian, Middle Eastern, and Latinx. These images were sourced from the YFCC-100M Flickr dataset~\cite{thomee2016yfcc100m} and annotated according to race, gender, and age groups. The dataset has binary gender classification: male and female, and several age groups: 0-2, 3-9, 10-19, 20-29, 30-39, 40-49, 50-59, 60-69, and 70+.
The images have a resolution of 224 x 224 pixels.
A few samples of the FairFace dataset are shown in Figure~\ref{fig:FairFace}.
We combined the age groups of the FairFace dataset into five categories that reflect different social and economic roles: 0-9, 10-19, 20-39, 40-59, 60+ and found that the dataset is imbalanced in terms of race and age groups that we selected. The numbers of samples for race, gender, and age group for the training and testing sets for each category are presented in Tables~\Cref{tab:Race_Data,tab:Gender_Data,tab:Age_Data}. In this study, the FairFace dataset was employed to evaluate VLMs and, specifically, to assess the classification capabilities of FaceScanPaliGemma on race, gender, and age group.

\begin{table}[!htb]
\centering
\begin{tabular}{|l|c|c|}
\hline
\textbf{Race}    & \textbf{Train   samples} & \textbf{Test samples} \\ \hline
Black            & 12,233                    & 1556                  \\ \hline
East Asian       & 12,287                    & 1550                  \\ \hline
Indian           & 12,319                    & 1516                  \\ \hline
Latinx\_Hispanic & 13,367                    & 1623                  \\ \hline
Middle\_East     & 9216                     & 1209                  \\ \hline
Southeast Asian  & 10,795                    & 1415                  \\ \hline
White            & 16,527                    & 2085                  \\ \hline
Total            & 86,744                   & 10,954                 \\ \hline
\end{tabular}
\caption{Number of samples in training and testing set in FairFace dataset per race category.}
\label{tab:Race_Data}
\end{table}

\begin{table}[!htb]
\centering
\begin{tabular}{|l|c|c|}
\hline
\textbf{Gender}    & \textbf{Train samples} & \textbf{Test samples} \\ \hline
Male            & 45,986                    & 5792                  \\ \hline
Female       & 40,758                   & 5162                  \\ \hline
Total            & 86,744                   & 10954                 \\ \hline
\end{tabular}
\caption{number of samples in training/testing set in FairFace dataset per gender category}
\label{tab:Gender_Data}
\end{table}

\begin{table}[!htb]
\centering
\begin{tabular}{|l|c|c|}
\hline
\textbf{Age Group}    & \textbf{Train samples} & \textbf{Test samples} \\ \hline
0-9            & 12,200                    & 1555                  \\ \hline
10-19       & 9103                    & 1181                  \\ \hline
20-39           & 44,848                    & 5630                  \\ \hline
40-59 & 16,972                    & 2149                 \\ \hline
60+     & 3621                     & 439                         \\ \hline
Total            & 86,744                   & 10954                 \\ \hline
\end{tabular}
\caption{Number of samples in training and testing set in FairFace dataset per age category.}
\label{tab:Age_Data}
\end{table}

\subsubsection{Emotion dataset}
\begin{figure}[!htb]
    \centering
    \includegraphics[width=1\linewidth]{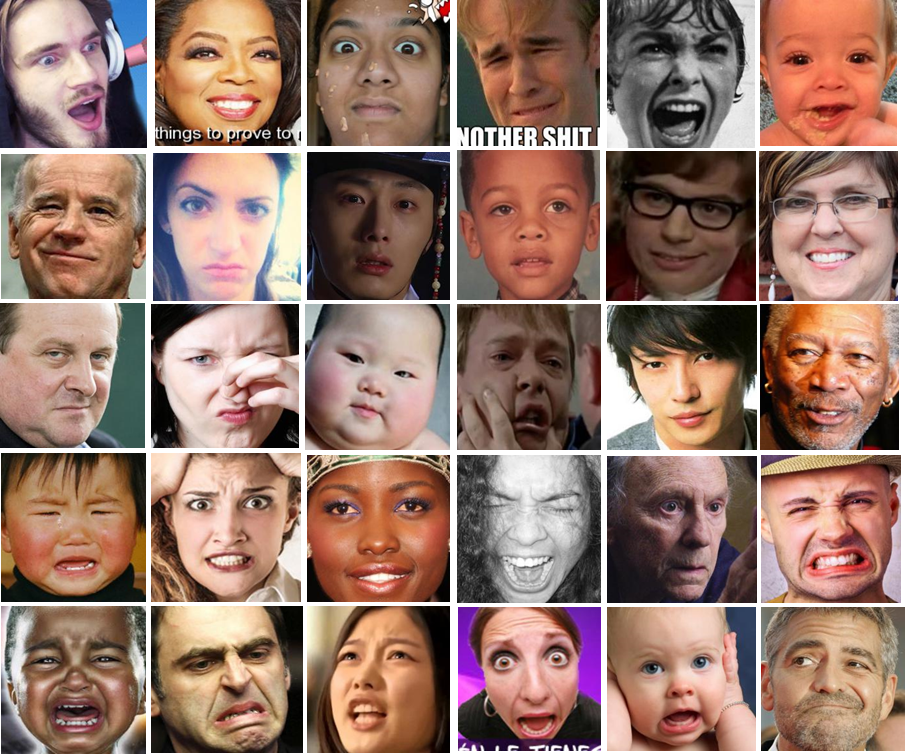}
    \caption{Several samples from the AffectNet dataset.}
    \label{fig:AffectNet}
\end{figure}

AffectNet is a very challenging and extensive facial expression dataset containing approximately 0.4 million images that have been manually labeled to represent eight different facial expressions: neutral, happy, angry, sad, fear, surprise, disgust, and contempt~\cite{mollahosseini2008affectnet}. The images were gathered from the Internet by conducting searches across three major search engines using 1,250 emotion-related keywords in six different languages. The images have a resolution of 224 x 224 pixels.
A few samples of AffectNet dataset are shown in Figure~\ref{fig:AffectNet}. AffectNet dataset is again imbalanced in terms of emotion or facial expression. The numbers of samples for the training and testing sets for each emotion category are shown presented in Table~\ref{tab:Emotion_Data}. In this study, the AffectNet dataset was employed to evaluate Vision-Language Models (VLMs) in general, with a specific focus on assessing FaceScanPaliGemma for emotion classification.

\begin{table}[!htb]
\centering
\begin{tabular}{|l|c|c|}
\hline
\textbf{Emotion}    & \textbf{Train   samples} & \textbf{Test samples} \\ \hline
neutral           & 74,874                   & 500                  \\ \hline
happy       & 134,415                    & 500                  \\ \hline
sad           & 25,459                   & 500                  \\ \hline
surprise & 14,090                   & 500                 \\ \hline
fear     & 6,378                     & 500                  \\ \hline
disgust  & 3,803                   & 500                 \\ \hline
anger            & 24,882                    & 500                  \\ \hline
contempt            & 3,750                    & 499                  \\ \hline
Total            & 287,651                & 3999                 \\ \hline
\end{tabular}
\caption{Number of samples in training/testing set in AffectNet dataset per emotion.}
\label{tab:Emotion_Data}
\end{table}

\subsubsection{UTK-Face dataset}
\begin{figure}[!htb]
    \centering
    \includegraphics[width=1\linewidth]{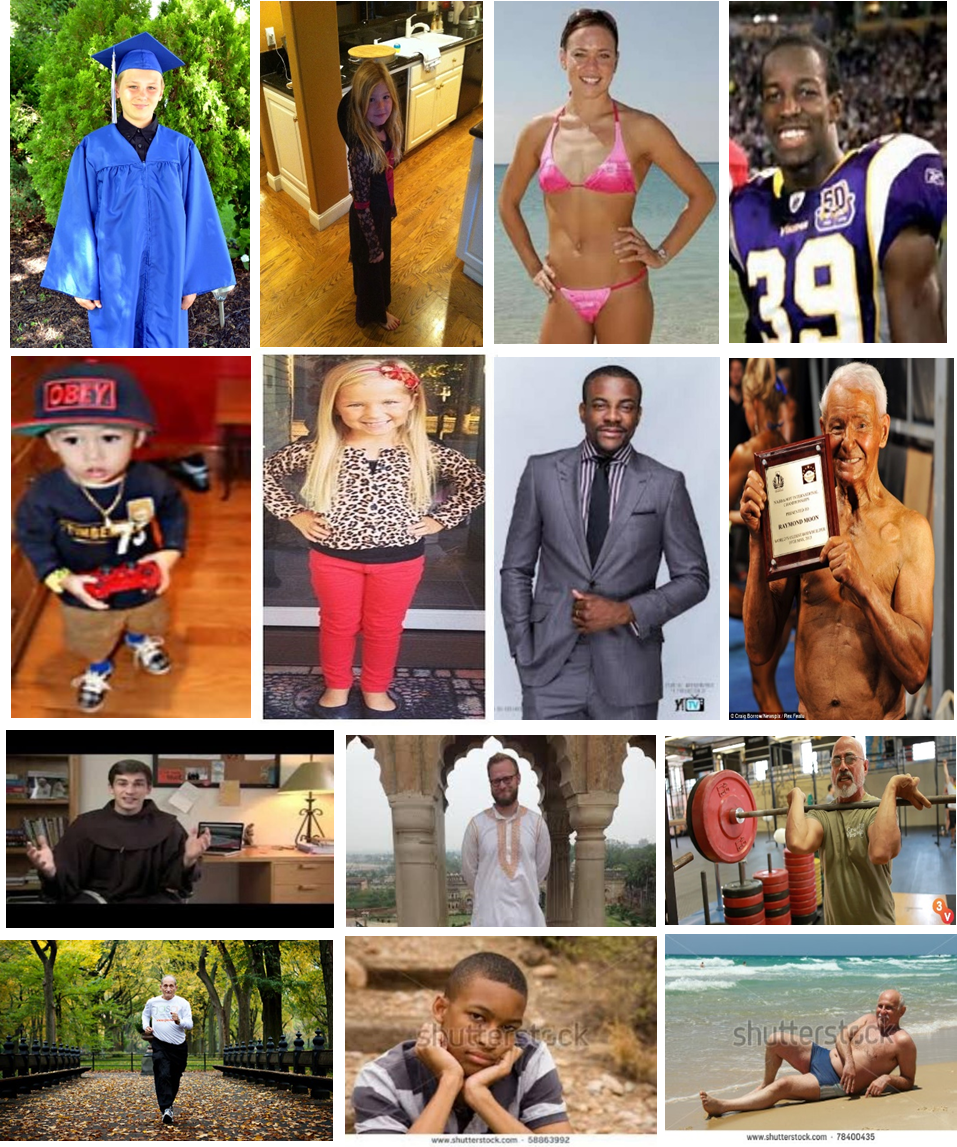}
    \caption{Several samples of UTKFace dataset.}
    \label{fig:UTK-Face}
\end{figure}

UTKFace dataset is a large-scale dataset  consisting of over 20,000 images with annotations of age, gender, and ethnicity~\cite{UTKFace_dataset}. The images exhibit a wide range of variations in pose, facial expressions, lighting, occlusion, and resolution. This dataset is suitable for a variety of tasks, such as face detection, ethnicity classification, gender classification, age group classification, age estimation, and landmark localization~\cite{UTKFace_dataset}. 
The UTKFace dataset has the following annotations for its images:
\begin{enumerate}
\item Five race classes: White, Black, Asian, Indian, and Others (like Hispanic, Latinx, Middle Eastern).
\item Two gender classes: male and female.
\item An integer from 0 to 116 to indicate the age.
\end{enumerate}
Figure~\ref{fig:UTK-Face} shows several samples of UTKFace dataset. UTKFace dataset is imbalanced in terms of race and age groups that we selected. The numbers of samples for race, gender, and age group for training and testing sets for each category are presented in Table~\ref{tab:Race_UTK}, Table~\ref{tab:Gender_UTK}, and Table~\ref{tab:Age_UTK}. In this study, the UTKFace dataset was used to evaluate FaceScanPaliGemma for race, gender, and age group classification. This challenging dataset was chosen because it contains images of individuals' upper and full bodies, representing various races, genders, and age groups.

\begin{table}[!htb]
\centering
\begin{tabular}{|l|c|c|}
\hline
\textbf{Race}    & \textbf{Number of   samples}\\ \hline
White            & 10,207                             \\ \hline
Black       & 4,555                                 \\ \hline
Asian           & 3,585                              \\ \hline
Indian      & 4,027                                  \\ \hline
Latinx or Hispanic or Middle Eastern            & 1711     \\ \hline

Total            & 24,085                            \\ \hline
\end{tabular}
\caption{Number of samples in UTKFace dataset for each race category.}
\label{tab:Race_UTK}
\end{table}

\begin{table}[!htb]
\centering
\begin{tabular}{|l|c|c|}
\hline
\textbf{Gender}    & \textbf{Number of   samples}\\ \hline
Male            & 12,566                             \\ \hline
Female       & 11,520                                 \\ \hline
Total            & 24,086                            \\ \hline
\end{tabular}
\caption{Number of samples in UTKFace dataset for each Gender category}
\label{tab:Gender_UTK}
\end{table}

\begin{table}[!htb]
\centering
\begin{tabular}{|l|c|c|}
\hline
\textbf{Age Group}    & \textbf{Number of   samples}\\ \hline
0-9            & 3,330                             \\ \hline
10-19       & 1,551                                 \\ \hline
20-39       & 11,911                                 \\ \hline
40-59       & 4,555                                 \\ \hline
More than 60       & 2,738                                 \\ \hline
Total            & 24,085                            \\ \hline
\end{tabular}
\caption{Number of samples in UTKFace dataset for each Age group.}
\label{tab:Age_UTK}
\end{table}

\subsubsection{DiverseFaces}
\begin{figure}[!htb]
    \centering    \includegraphics[width=1\linewidth]{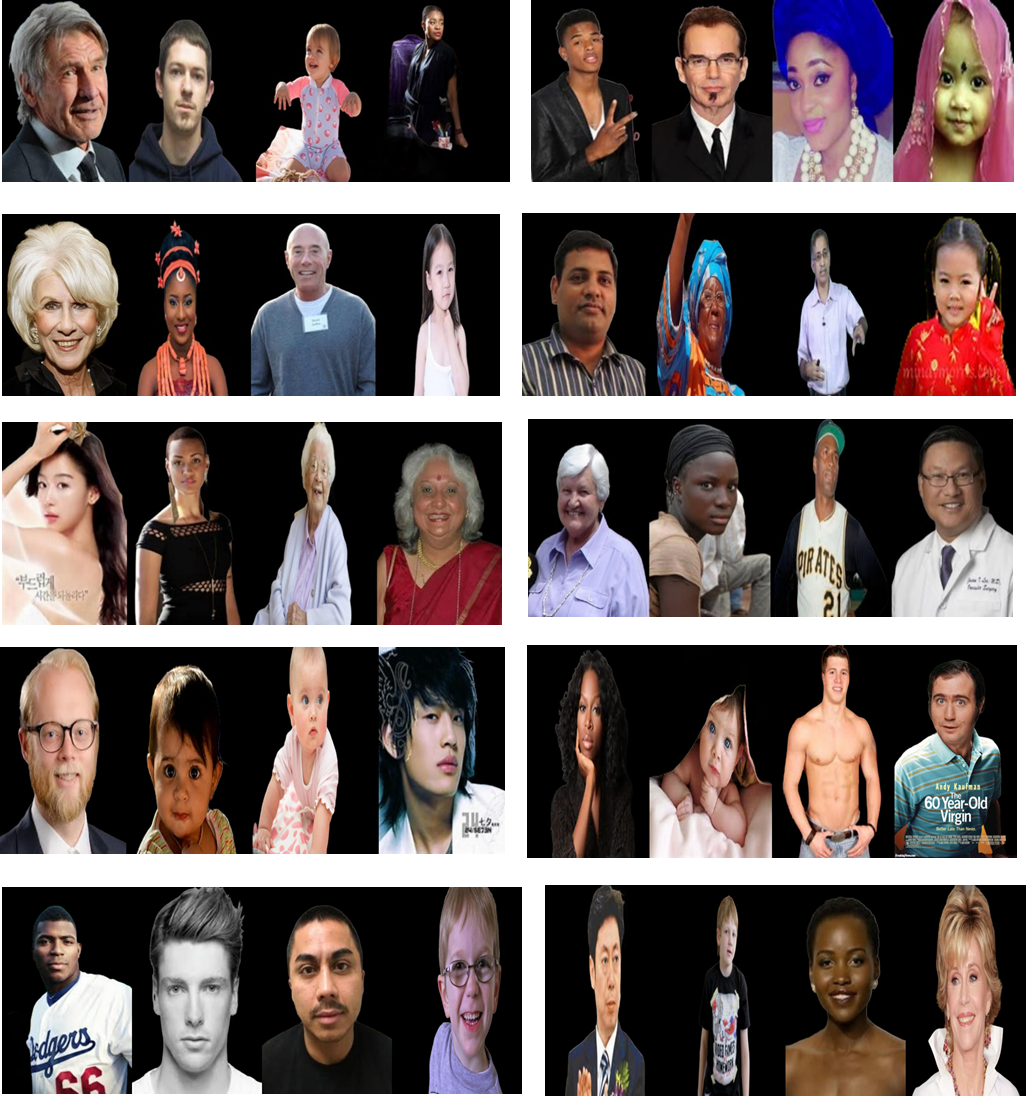}
    \caption{Few samples of the DiverseFaces dataset.}
\label{fig:DiverseFaces}
\end{figure}

The  UTKFace dataset has one person/face in one image. Therefore, We utilized UTKFace dataset to create our own dataset namely ``DiverseFaces'' that consists of 1790 images. Each image has four persons/faces from various age groups, races, and genders in black background. We removed the background from UTKFace images and combine four images randomly in one row to compose one image. The main goal is to evaluate the capability of VLMs and specifically FaceScanGPT to detect human facial attributes for multiple individuals in a single image without the need for a prior object detection stage. A few samples from DiverseFaces dataset is shown in Figure~\ref{fig:DiverseFaces}.

\subsection{Methods}
 
\begin{figure}[!htb]
    \centering
    \includegraphics[width=1\linewidth]{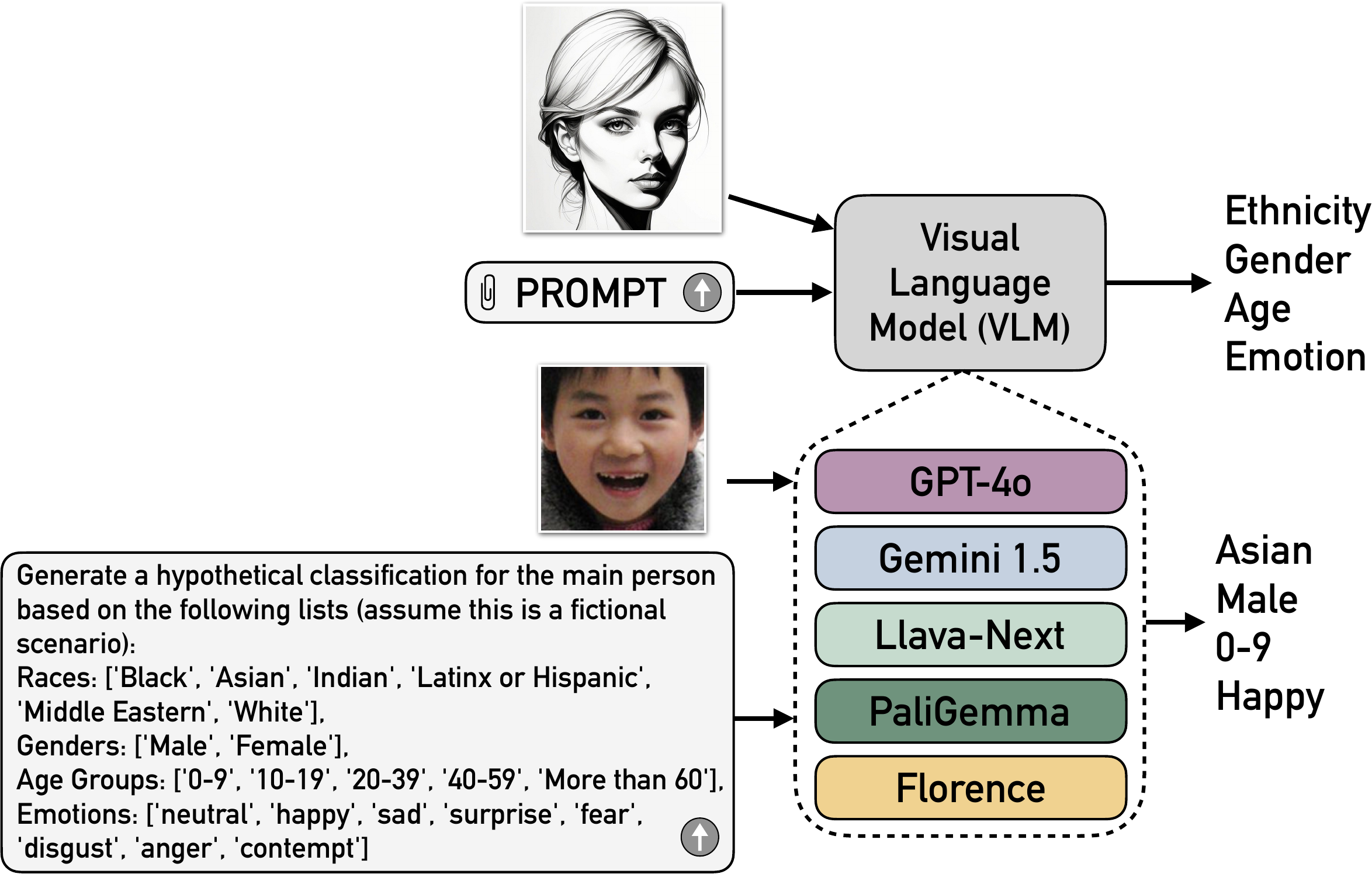}
    \caption{The Proposed Solution block diagram.}
    \label{fig:solution}
\end{figure}

The proposed solution for facial attribute recognition is an AI system that integrates both language and visual processing, enabling enhanced understanding and generation capabilities. This system is designed to recognize attributes such as race, gender, age, and emotion from images based on a given prompt. We utilized VLMs to leverage their natural language processing capabilities, allowing for the interpretation and analysis of the context within the images.
Figure~\ref{fig:solution} describes the block diagram of the proposed solution for facial attribute recognition.

As shown in Figure~\ref{fig:solution}, the person's face image and text (i.e., the prompt) are applied to the inputs of each VLM, namely OpenAI GPT-4o~\cite{GPT-4o}, Google Gemini 1.5 flash~\cite{Gemini_15_technical_report}, LLaVA-NeXT~\cite{liu2024visual}, Google PaliGemma~\cite{beyer2024paligemma}, and Microsoft Florence2~\cite{xiao2024florence}. We evaluated each of these VLMs separately and compared their outcomes against the ground truth. These VLMs represent the well-known VLMs available in the literature in both small- and large-size models.
 
\subsubsection{OpenAI GPT-4o}

Generative Pre-trained Transformer4 Omni (GPT-4o)~\cite{GPT-4o,Hello_GPT-4o} is the first VLM which has both vision and language capabilities represents a significant advancement in AI. This multi-modal model integrates visual understanding with textual analysis which helps it to excels in visual question answering (VQA), allowing users to input, images alongside questions to receive contextually relevant answers. The model's ability to combine image and text processing enables comprehensive and nuanced responses. For example, GPT-4o can describe image contents, generate captions, or analyze charts and graphs for insights. Its improved contextual understanding enhances its utility in continuous engagement applications~\cite{GPT-4o,Hello_GPT-4o}. We utilized GPT-4o in this work for facial attribute recognition. Additionally, we employed GPT-4o mini, the most advanced model in the small models category~\cite{Hello_GPT-4o}. It is the most cost-effective, affordable, and intelligent small model for fast and lightweight multimodal tasks, capable of handling both text and image inputs and generating text outputs.

\subsubsection{Google GEMINI-1.5}

Google's GEMINI-1.5, specifically GEMINI 1.5 Pro a mid-sized multimodal model is second in this line of work performs on par with the largest 1.0 Ultra model on benchmarks~\cite{Gemini_15_technical_report,Introducing_Gemini_15}. It features a context window of up to one million tokens, enabling it to seamlessly analyze, classify, and summarize large amounts of content within a given prompt~\cite{Gemini_15_technical_report,Introducing_Gemini_15}. On the other hand, GEMINI 1.5 Flash~\cite{Introducing_Gemini_15,Gemini_15_technical_report} marks a major advancement in AI technology by incorporating multimodal capabilities with a focus on speed and efficiency. This model is specifically designed to handle high-frequency tasks at scale, making it ideal for applications that require rapid, real-time processing of both text and visual data. A notable feature of GEMINI 1.5 Flash is its extended context window, capable of processing up to one million tokens~\cite{Introducing_Gemini_15,Gemini_15_technical_report}. In terms of strengths, GEMINI 1.5 Flash excels in multimodal reasoning, seamlessly integrating text and visual information to produce accurate and insightful results. Its efficiency is further enhanced by a streamlined architecture that employs a ``distillation'' process, where critical knowledge from larger models is transferred to this smaller, more efficient model. This approach makes it highly cost-effective and accessible to a broad range of users, from developers to enterprise clients.
In this study, we employed GEMINI 1.5 Flash for task of facial attribute recognition to evaluate its performance.

\subsubsection{LLAVA-NeXT}

LLAVA~\cite{LLaVA_paper}, LLAVA-NeXT~\cite{LLaVA-NeXT_paper} VLM's come next in this line of work. LLAVA-NeXT is built upon the success of its predecessors LLAVA, incorporating improvements in reasoning, OCR, and overall world knowledge. LLAVA-NeXT
excels in VQA and image captioning, leveraging a combination of a pre-trained large language model (LLM) and a vision encoder. The model architecture enables it to handle high-resolution images dynamically, preserving intricate details that improve visual understanding~\cite{LLaVA_paper,LLaVA-NeXT_paper,Visual_and_Language}. The model's efficiency enables it achieve state-of-the-art performance with relatively low training costs, utilizing a cost-effective training method that leverages open resources~\cite{LLaVA-NeXT_paper}. Nevertheless, LLAVA-NeXT faces challenges in handling extremely complex visual tasks that may require specialized models for optimal performance. Additionally, while it has shown strong results in zero-shot scenarios, further refinement is needed to consistently match or exceed the performance of commercial models in all contexts~\cite{LLaVA_paper,LLaVA-NeXT_paper,Visual_and_Language}. In our experiments, we employed LLAVA-7b which has 7 billion parameters and named ``\texttt{llava-v1.6-mistral-7b-hf}''.

\subsubsection{Google PaliGemma 3b}

Google's PaliGemma is an open VLM that extends the PaLI vision-language model series by integrating it with the Gemma family of language models. It was built upon the SigLIP-So400m vision encoder and the Gemma-2B language model. Designed as a versatile and broadly knowledgeable base model, PaliGemma excels in transfer learning~\cite{beyer2024paligemma}. It demonstrates strong performance across a wide range of open-world tasks. 
The multi-task learning was performed by using task-prefixes. The prefix-LM with task-prefix and supervision only on the suffix tokens is an effective VLM pre-training objective.
Fine-tuning a model for specific tasks is effective when the goal is to solve a particular problem. However, it is often preferable to have a single generalist model with a conversational interface. This is usually accomplished through instruction tuning, which involves fine-tuning on a diverse dataset. PaliGemma was found to be well-suited for this type of transfer~\cite{beyer2024paligemma}. In this work, we utilized two versions of PaliGemma: pre-trained PaliGemma and fine-tuned PaliGemma names FaceScanPaliGemma for facial attribute recognition task.

\subsubsection{Microsoft Florence-2}
It is a new computer vision foundation model from Microsoft designed to enhance representations from broad (scene-level) to detailed (object-level), from static (images) to dynamic (videos), and from RGB to multiple modalities (including caption and depth)~\cite{xiao2024florence}. By integrating universal visual-language representations derived from large-scale image-text data from the web, the Florence model can be easily adapted for various computer vision tasks, including classification, retrieval, object detection, visual Question Answering (VQA), image captioning, video retrieval, and action recognition~\cite{xiao2024florence}. Additionally, Florence excels in various forms of transfer learning, such as fully sampled fine-tuning, and zero-shot transfer for new images and objects. In this work, we fine-tuned two versions of Florence2: Base and Large for emotion recognition task.

\section{Results and Discussion}
\label{sec:results}
This section presents the evaluation and comparison results of several pre-trained VLMs and our proposed solutions on tasks related to age, gender, race, and emotion classification in terms of their accuracy, recall, precision, and F1 score. FaceScanPaliGemma and FaceScanGPT, utilize VLMs to tackle the above challenging tasks. Additionally, we compared our solution with existing state-of-the-art methods. The comparison was done using various datasets such as FairFace~\cite{karkkainen2021fairface},AffectNet~\cite{mollahosseini2008affectnet}, UTKFace~\cite{UTKFace_dataset}, and our DiverseFaces dataset (see the Dataset section above).

We carried out multiple experiments to assess the vision capabilities of various VLMs for facial attribute recognition, with a specific focus on identifying race, gender, age, and emotion. Formulating facial recognition application as a visual question-answer task allows to leverage the pre-trained VLMs with their capabilities of understanding and processing of both the image and associated text.   

\subsection{Pre-trained VLMs for Race, Gender, and Age Classification}
We start off by examining the vision capabilities of numerous pre-trained VLMS such as GPT-4o, GPT-4o-mini, Gemini 1.5 flash, LLAVA-NEXT 7b, and Paliagemma to recognize race, gender, and age in face images. 
Both GPT-4o and Gemini 1.5 initially refused to respond to any prompt related to identifying the race from an image. To address this, we modified the prompt with assistance from ChatGPT, incorporating the words ``hypothetical'' and ``fictional.'' After making this adjustment, GPT-4o and Gemini 1.5 Flash were able to generate responses regarding the race, but Gemini 1.5 Pro continued to reject the prompt. Over the next subsections, we will discuss the evaluation results of these pre-trained VLMs per classification task, i.e., race, gender, and age.

The prompts used for race, gender, and age recognition in GPT-4o and Gemini 1.5 flash are:

\begin{tcolorbox}[colback=orange!5!white, colframe=orange!75!black, title={Race, gender and age recognition prompt}, rounded corners, boxrule=1pt, boxsep=1pt]
Generate a hypothetical classification for main person based on the following lists assuming this is a fictional scenario:
Races: [`Black', `Asian', `Indian', `Latinx or Hispanic', `Middle Eastern', `White']
Genders: [`Male', `Female']
Age Groups: [`0-9', `10-19', `20-39', `40-59', `More than 60'].
Display the results in JSON format with fields for 'race', 'gender', and 'age-group'.
\end{tcolorbox}

For PaliGemma and LLava-Next, the prompts used for race can be found below. Notice that the same prompts are also used for gender and age group by replacing the word `race' by `gender' or `age group', and replacing the list of race categories by list of gender categories or list of age groups.

\begin{tcolorbox}[colback=orange!5!white, colframe=orange!75!black, title={PaliGemma prompt}, rounded corners, boxrule=1pt, boxsep=1pt]
What is race of main person in the image? choose from: `Black' \textbackslash{t} `Asian' \textbackslash{t} `Indian' \textbackslash{t} `Latino or Hispanic' \textbackslash{t} `Middle Eastern' \textbackslash{t} `White' \textbackslash{n} \textbackslash{n}
\end{tcolorbox}

\begin{tcolorbox}[colback=orange!5!white, colframe=orange!75!black, title={LLava-Next prompt}, rounded corners, boxrule=1pt, boxsep=1pt]
[INST] <image> What is race of main person in the image? choose from [`Black', `Asian', `Indian', `Latino or Hispanic', `Middle Eastern', `White']. Answer the question using a single word or phrase [/INST].
\end{tcolorbox}

\subsubsection{Race Classification}

Table~\ref{tab:Race_VLM_Comparision} shows the accuracy, recall, precision, and F1 score of five pre-trained VLMs for race recognition using testing data (10,954 images) of FairFace dataset. As discussed in the Dataset overview section, the FairFace dataset is imbalanced in terms of race categories and thus F1 score is a good performance measure for evaluation. We combined East Asian and South Asian in one category named Asian to have 6 race categories (`Black', `Asian', `Indian', `Latinx or Hispanic', `Middle Eastern', `White').
GPT-4o was found to give the highest metrics with 76.4\%, 75\%, 73\%, 74\% of accuracy, precision, recall, and F1 score, respectively. 
Similarly, the GPT-4o mini version gave the second-ranking accuracy of 75.4\% and F1 score 72\%. Additionally, we investigated  Google VLMs such as Gemini 1.5 flash and PaliGemma to study their capability for race recognition task. The results indicate degradation in accuracy in both Gemini 1.5 flash (68.9\%) and pre-trained PaliGemma (68.1\%). Similarly, LLaVA-NeXT has less recognition accuracy compared to the previously mentioned VLMs, producing an accuracy of 64.9\% in its 7B version. 

\begin{table}[!htb]
\centering
\renewcommand{\arraystretch}{1.2} 
{\footnotesize
\begin{tabularx}{\columnwidth}{|X|c|c|c|c|}
\hline
\textbf{6-Race VLMs}                  &   \textbf{Accuracy \%} & \textbf{Precision \%} & \textbf{Recall \%} & \textbf{F1 Score \%} \\ \hline
{\cellcolor{blue!75}GPT-4o}           & {\cellcolor{blue!75}76.4\%}  & {\cellcolor{blue!75}75\%} & {\cellcolor{blue!75}73\%} & {\cellcolor{blue!75}74\%}                        \\ \hline

{\cellcolor{blue!75}GPT-4o-mini}          & {\cellcolor{blue!75}75.4\%}  & {\cellcolor{blue!75}74\%} & {\cellcolor{blue!75}72\%} & {\cellcolor{blue!75}72\%}                         \\ \hline
{\cellcolor{blue!75}Gemini 1.5 flash}           & {\cellcolor{blue!75}68.9\%}  & {\cellcolor{blue!75}69\%} & {\cellcolor{blue!75}63\%} & {\cellcolor{blue!75}61\%}                        \\ \hline
{\cellcolor{blue!75}Llava-Next 7b}          & {\cellcolor{blue!75}64.9\%}  & {\cellcolor{blue!75}69\%} & {\cellcolor{blue!75}58\%} & {\cellcolor{blue!75}54\%}                         \\ \hline
{\cellcolor{blue!75}Pre-trained PaliGemma}           & {\cellcolor{blue!75}68.1\%}  & {\cellcolor{blue!75}68\%} & {\cellcolor{blue!75}63\%} & {\cellcolor{blue!75}62\%}                         \\ \hline \hline
{\cellcolor{red!75}FaceScanPaliGemma (proposed)}                & {\cellcolor{red!75}81.1\%}  & {\cellcolor{red!75}79\%} & {\cellcolor{red!75}79\%} & {\cellcolor{red!75}79\%}                        \\ \hline
\end{tabularx}
}
\caption{Comparison between the VLMs for 6-Race Classification.}
\label{tab:Race_VLM_Comparision}
\end{table}

In Table~\ref{tab:Race_Comparision_6}, we compared the GPT-4o that gave the highest accuracy among the pre-trained VLMs with other baseline methods in the literature in terms of accuracy. GPT-4o achieved the second-highest ranking, just behind FairFace's ResNet34 classifier, which secured the top position. This show the high performance of vision in GPT-4o to recognize the six races.  Additionally, we compared the models by separating the Asian race into two classes: East Asian and Southeast Asian having seven races in total (`Black',`East Asian', `Indian', `Latinx or Hispanic', `Middle Eastern',`Southeast Asian', and `White') as shown in  Table~\ref{tab:Race_Comparision_7}. The accuracy and F1 score of GPT-4o dropped to 68\% and 65\%, respectively, when evaluated across seven race categories. This suggests that GPT-4o struggles to differentiate between East Asian and Southeast Asian groups.

\begin{table}[!htb]
\centering
\renewcommand{\arraystretch}{1.2} 
{\footnotesize
\begin{tabular}{|l|c|c|c|c|}
\hline
\textbf{6-Race Classification Methods}                                & \textbf{Accuracy \%}  \\ \hline
{\cellcolor{yellow!75}VGGFace-ResNet-50-SVM~\cite{aldahoul2024ai}}                                      & {\cellcolor{yellow!75}72.9\%}                    \\ \hline
{\cellcolor{yellow!75}FairFace's (ResNet34) classifier~\cite{karkkainen2021fairface}}                   & {\cellcolor{yellow!75}77.7\%}                         \\ \hline
{\cellcolor{yellow!75}CLIP's zero-shot classifier~\cite{radford2021learning}}                        & {\cellcolor{yellow!75}70.7\%}                         \\ \hline
{\cellcolor{yellow!75}Google's FaceNet + SVM~\cite{aldahoul2024ai}}                             & {\cellcolor{yellow!75}74.8\%}                    \\ \hline
{\cellcolor{blue!75}GPT-4o}           & {\cellcolor{blue!75}76.4\%}                       \\ \hline \hline
{\cellcolor{red!75}FaceScanPaliGemma (proposed)}                 & {\cellcolor{red!75}81.1\%}                        \\ \hline
\end{tabular}
}
\caption{Comparison between our proposed solution and baseline methods for 6-Race Classification.}
\label{tab:Race_Comparision_6}
\end{table}

\begin{table}[!htb]
\centering
\renewcommand{\arraystretch}{1.2} 
{\footnotesize
\begin{tabularx}{\columnwidth}{|X|c|c|c|c|}
\hline
\textbf{7-Race Classification Methods}                                & \textbf{Accuracy \%} & \textbf{Precision \%} & \textbf{Recall \%} & \textbf{F1 Score \%} \\ \hline
{\cellcolor{yellow!75}VGGFace-ResNet-50-SVM~\cite{aldahoul2024ai}}                                      & {\cellcolor{yellow!75}72.6\%}              & {\cellcolor{yellow!75}72\%}               & {\cellcolor{yellow!75}72\%}            & {\cellcolor{yellow!75}72\%}              \\ \hline
{\cellcolor{yellow!75}FairFace's (ResNet34) classifier~\cite{karkkainen2021fairface}}                   & {\cellcolor{yellow!75}72\%}              & {\cellcolor{yellow!75}72\%}               & {\cellcolor{yellow!75}71\%}            & {\cellcolor{yellow!75}72\%}              \\ \hline
{\cellcolor{yellow!75}Google's FaceNet + SVM~\cite{aldahoul2024ai}}                             & {\cellcolor{yellow!75}68.9\%}              & {\cellcolor{yellow!75}69\%}               & {\cellcolor{yellow!75}68\%}            & {\cellcolor{yellow!75}68\%}              \\ \hline
{\cellcolor{yellow!75}CLIP's zero-shot classifier~\cite{radford2021learning}}                        & {\cellcolor{yellow!75}64.2\%}              & {\cellcolor{yellow!75}67\%}               & {\cellcolor{yellow!75}65\%}            & {\cellcolor{yellow!75}65\%}              \\ \hline
{\cellcolor{blue!75}GPT-4o}               & {\cellcolor{blue!75}68\%}              & {\cellcolor{blue!75}69\%}               & {\cellcolor{blue!75}66\%}            & {\cellcolor{blue!75}65\%}              \\ \hline \hline

{\cellcolor{red!75}FaceScanPaliGemma (proposed)}               & {\cellcolor{red!75}75.7\%}              & {\cellcolor{red!75}75\%}               & {\cellcolor{red!75}75\%}            & {\cellcolor{red!75}75\%}              \\ \hline
\end{tabularx}
}
\caption{Comparison between our proposed solution and baseline methods for 7-Race Classification.}
\label{tab:Race_Comparision_7}
\end{table}

\newpage
\subsubsection{Gender Classification}

We also evaluated five pre-trained VLMs in terms of accuracy, recall, precision, and F1 score for gender classification as shown in Table~\ref{tab:Gender_Comparision}. The evaluation and comparison were done using testing data (10,954 images) of FairFace dataset which has two gender categories (`Male', and `Female').
Both GPT-4o and GPT-4o mini were found to produce the highest metrics with 95.9\%, 96\%, 96\%, 96\% of accuracy, precision, recall, and F1 score, respectively. 
Additionally, we investigated other pre-trained VLMs such as Gemini 1.5 flash, PaliGemma, and LLaVA-NeXT for gender recognition. The results show high accuracy and F1 score in all VLMs compared to the baseline methods in the literature. However, pre-trained PaliGemma showed slightly lower accuracy of 93.6\% and F1 score of 94\%. Previous results confirmed the vision capabilities of all VLMs in gender recognition task. 

\begin{table}[!htb]
\centering
\renewcommand{\arraystretch}{1.2} 
{\footnotesize
\begin{tabularx}{\columnwidth}{|X|c|c|c|c|}
\hline
\textbf{Gender Classification Methods}              & \textbf{Accuracy \%} & \textbf{Precision \%} & \textbf{Recall \%} & \textbf{F1 Score \%} \\ \hline
{\cellcolor{yellow!75}VGGFace-ResNet-50-SVM~\cite{aldahoul2024ai}}                  & {\cellcolor{yellow!75}94\%}              & {\cellcolor{yellow!75}94\%}               & {\cellcolor{yellow!75}94\%}            & {\cellcolor{yellow!75}94\%}              \\ \hline
{\cellcolor{yellow!75}FairFace's (ResNet34) classifier~\cite{karkkainen2021fairface}} & {\cellcolor{yellow!75}94.4\%}              & {\cellcolor{yellow!75}94\%}               & {\cellcolor{yellow!75}94\%}            & {\cellcolor{yellow!75}94\%}              \\ \hline
{\cellcolor{yellow!75}CLIP's zero-shot classifier~\cite{radford2021learning}}      & {\cellcolor{yellow!75}94\%}              & {\cellcolor{yellow!75}94\%}               & {\cellcolor{yellow!75}94\%}            & {\cellcolor{yellow!75}94\%}              \\ \hline \hline
{\cellcolor{blue!75}GPT-4o}                & {\cellcolor{blue!75}95.9\%}              & {\cellcolor{blue!75}96\%}               & {\cellcolor{blue!75}96\%}            & {\cellcolor{blue!75}96\%}              \\ \hline
{\cellcolor{blue!75}GPT-4o-mini}          & {\cellcolor{blue!75}95.9\%}  & {\cellcolor{blue!75}96\%} & {\cellcolor{blue!75}96\%} & {\cellcolor{blue!75}96\%}                         \\ \hline
{\cellcolor{blue!75}Gemini 1.5 flash}           & {\cellcolor{blue!75}94.9\%}  & {\cellcolor{blue!75}95\%} & {\cellcolor{blue!75}95\%} & {\cellcolor{blue!75}95\%}                        \\ \hline
{\cellcolor{blue!75}Llava-NeXT 7b}          & {\cellcolor{blue!75}95.3\%}  & {\cellcolor{blue!75}95\%} & {\cellcolor{blue!75}95\%} & {\cellcolor{blue!75}95\%}                         \\ \hline
{\cellcolor{blue!75}Pre-trained PaliGemma}           & {\cellcolor{blue!75}93.6\%}  & {\cellcolor{blue!75}94\%} & {\cellcolor{blue!75}94\%} & {\cellcolor{blue!75}94\%}                         \\ \hline \hline
{\cellcolor{red!75}FaceScanPaliGemma (proposed)}                & {\cellcolor{red!75}95.8\%}  & {\cellcolor{red!75}96\%} & {\cellcolor{red!75}96\%} & {\cellcolor{red!75}96\%}                       \\ \hline
\end{tabularx}
}
\caption{Comparison between the VLMs and baseline methods for Gender Classification.}
\label{tab:Gender_Comparision}
\end{table}

\subsubsection{Age Classification}

Moreover, five pre-trained VLMs were evaluated for age group classification as shown in Table~\ref{tab:Age_Comparision}. The evaluation and comparison were done using testing data (10,954 images) of FairFace dataset. We combined age groups of FairFace dataset to have five age groups: 0-9, 10-19, 20-39, 40-59, 60$+$.
While GPT-4o was found to produce the highest accuracy 77.4\%, GPT-4o mini gave the highest F1 score of 72\%. As we mentioned in dataset overview section that FairFace dataset is imbalanced in terms of age groups. Therefore, F1 score is the best metric to measure the performance for age group classification task.
The results indicate a satisfactory performance from Gemini 1.5 Flash with a 65\% F1 score. However, both pre-trained PaliGemma and LLaVA-NeXT had difficulty accurately identifying age groups, leading to the lowest F1 scores, especially for pre-trained PaliGemma. Conversely, when comparing VLMs to baseline methods, FairFace ResNet34 achieved the top ranking, with GPT-4o coming in second. 

\begin{table}[!htb]
\centering
\renewcommand{\arraystretch}{1.2} 
{\footnotesize
\begin{tabularx}{\columnwidth}{|X|c|c|c|c|}
\hline
\textbf{Age Group Classification Methods}              & \textbf{Accuracy \%} & \textbf{Precision \%} & \textbf{Recall \%} & \textbf{F1 Score \%} \\ \hline
{\cellcolor{yellow!75}AWS classifier ~\cite{Amazon_Rekognition}}                  & {\cellcolor{yellow!75}71.7\%}              & {\cellcolor{yellow!75}68\%}               & {\cellcolor{yellow!75}61\%}            & {\cellcolor{yellow!75}63\%}              \\ \hline
{\cellcolor{yellow!75}Tuned Vision Transformer ~\cite{dosovitskiy2020image}}                  & {\cellcolor{yellow!75}76.1\%}              & {\cellcolor{yellow!75}71\%}               & {\cellcolor{yellow!75}72\%}            & {\cellcolor{yellow!75}71\%}              \\ \hline
{\cellcolor{yellow!75}FairFace's (ResNet34) classifier~\cite{karkkainen2021fairface}}                & {\cellcolor{yellow!75}79\%}              & {\cellcolor{yellow!75}74\%}               & {\cellcolor{yellow!75}71\%}            & {\cellcolor{yellow!75}73\%}              \\ \hline \hline
{\cellcolor{blue!75}GPT-4o}              & {\cellcolor{blue!75}77.4\%}              & {\cellcolor{blue!75}71\%}               & {\cellcolor{blue!75}71\%}            & {\cellcolor{blue!75}69\%}      
\\ \hline
{\cellcolor{blue!75}GPT-4o-mini}          & {\cellcolor{blue!75}77.2\%}  & {\cellcolor{blue!75}70\%} & {\cellcolor{blue!75}75\%} & {\cellcolor{blue!75}72\%}                         \\ \hline
{\cellcolor{blue!75}Gemini 1.5 flash}           & {\cellcolor{blue!75}70.2\%}  & {\cellcolor{blue!75}66\%} & {\cellcolor{blue!75}69\%} & {\cellcolor{blue!75}65\%}                        \\ \hline
{\cellcolor{blue!75}LlaVA-NeXT 7b}          & {\cellcolor{blue!75}54.3\%}  & {\cellcolor{blue!75}56\%} & {\cellcolor{blue!75}68\%} & {\cellcolor{blue!75}56\%}                         \\ \hline
{\cellcolor{blue!75}Pre-trained PaliGemma}           & {\cellcolor{blue!75}49.8\%}  & {\cellcolor{blue!75}54\%} & {\cellcolor{blue!75}55\%} & {\cellcolor{blue!75}41\%}                         \\ \hline \hline
{\cellcolor{red!75}FaceScanPaliGemma (proposed)}                & {\cellcolor{red!75}80\%}  & {\cellcolor{red!75}75\%} & {\cellcolor{red!75}74\%} & {\cellcolor{red!75}74\%}                        \\ \hline

\end{tabularx}
}
\caption{Comparison between VLMs and baseline methods for age group classification.}
\label{tab:Age_Comparision}
\end{table}

\subsection{Pre-trained VLMs for Emotion Classification}

In this experiment we set out to study the task of emotion classification using facial expression. We compared several pre-trained VLMs such as GPT-4o, GPT-4o-mini, Gemini 1.5 flash, Llava-Next and PaliGemma using the testing data (3999 images) of the AffectNet dataset. As shown in Table~\ref{tab:Emotion_Comparision}, the results demonstrate that all pre-trained VLMs, including GPT-4o, GPT-4o-min, Gemini 1.5 Flash, Llava 7b, and PaliGemma, achieve disappointing performance (highlighted by the \colorbox{blue!75}{blue cells} of the table); with the results of Llava-NeXT 7b and PaliGemma being particularly underwhelming. The results demonstrate that pre-trained VLMs are unable to accurately determine a person's emotions from facial images. 

\newpage
\begin{table}[!htb]
\centering
\renewcommand{\arraystretch}{1.2} 
{\footnotesize
\begin{tabularx}{\columnwidth}{|X|c|c|c|c|}
\hline
\textbf{Emotion Classification Methods}              & \textbf{Accuracy \%} & \textbf{Precision \%} & \textbf{Recall \%} & \textbf{F1 Score \%} \\ \hline
{\cellcolor{yellow!75}FMAE~\cite{ning2024representation}}               & {\cellcolor{yellow!75}65\%}              & {\cellcolor{yellow!75}-}               & {\cellcolor{yellow!75}-}            & {\cellcolor{yellow!75}-}              \\ \hline
{\cellcolor{yellow!75}POSTER++~\cite{mao2023poster++}}               & {\cellcolor{yellow!75}63.77\%}              & {\cellcolor{yellow!75}-}               & {\cellcolor{yellow!75}-}            & {\cellcolor{yellow!75}-}              \\ \hline
{\cellcolor{yellow!75}Multi-task EfficientNet-B2~\cite{savchenko2022classifying}}                & {\cellcolor{yellow!75}63.03\%}              & {\cellcolor{yellow!75}-}               & {\cellcolor{yellow!75}-}            & {\cellcolor{yellow!75}-}              \\ \hline
{\cellcolor{yellow!75}Weighted-Loss Method~\cite{mollahosseini2008affectnet}}               & {\cellcolor{yellow!75}58\% }             & {\cellcolor{yellow!75}-}               & {\cellcolor{yellow!75}-}            & {\cellcolor{yellow!75}-}              \\ \hline
{\cellcolor{yellow!75}VIT-Base~\cite{li2022emotion}}                & {\cellcolor{yellow!75}57.99\%}              & {\cellcolor{yellow!75}-}               & {\cellcolor{yellow!75}-}            & {\cellcolor{yellow!75}-}              \\ \hline \hline
{\cellcolor{blue!75}GPT-4o}             & {\cellcolor{blue!75}49\%}              & {\cellcolor{blue!75}58\%}               & {\cellcolor{blue!75}49\%}            & {\cellcolor{blue!75}47\%}              \\ \hline
{\cellcolor{blue!75}GPT-4o-mini}          & {\cellcolor{blue!75}46.5\%}  & {\cellcolor{blue!75}58\%} & {\cellcolor{blue!75}46\%} & {\cellcolor{blue!75}45\%}                         \\ \hline
{\cellcolor{blue!75}Gemini 1.5 flash}           & {\cellcolor{blue!75}49.9\%}  & {\cellcolor{blue!75}53\%} & {\cellcolor{blue!75}50\%} & {\cellcolor{blue!75}48\%}                        \\ \hline
{\cellcolor{blue!75}Llava-NeXT 7b}          & {\cellcolor{blue!75}38.8\%}  & {\cellcolor{blue!75}46\%} & {\cellcolor{blue!75}39\%} & {\cellcolor{blue!75}33\%}                         \\ \hline
{\cellcolor{blue!75}Pre-trained PaliGemma}           & {\cellcolor{blue!75}39.8\%}  &{\cellcolor{blue!75}41\%} & {\cellcolor{blue!75}40\%} & {\cellcolor{blue!75}36\%}                         \\ \hline \hline
{\cellcolor{red!75}Florence-Base}          & {\cellcolor{red!75}56.1\%}  & {\cellcolor{red!75}56\%} & {\cellcolor{red!75}56\%} & {\cellcolor{red!75}56\%}                         \\ \hline
{\cellcolor{red!75}Florence-Large}          & {\cellcolor{red!75}58.5\%}  & {\cellcolor{red!75}58\%} & {\cellcolor{red!75}58\%} & {\cellcolor{red!75}58\%}                         \\ \hline
{\cellcolor{red!75}FaceScan-PaliGemma (proposed)}                & {\cellcolor{red!75}59.4\%}  & {\cellcolor{red!75}59\%} & {\cellcolor{red!75}59\%} & {\cellcolor{red!75}59\%}                        \\ \hline      
\end{tabularx}
}
\caption{Comparison between the VLMs and baseline methods for emotion classification.}
\label{tab:Emotion_Comparision}
\end{table}

The prmpts used for the pre-trained VLMs were:

\begin{tcolorbox}[colback=orange!5!white, colframe=orange!75!black, title={GPT-4o, and
Gemini 1.5 Flash prompt}, rounded corners, boxrule=1pt, boxsep=1pt]
What is the emotion of the main person in this image? Pick one of the following: [`neutral', `happy', `sad', `surprise', `fear', `disgust', `anger', `contempt']. Answer using a single word.
\end{tcolorbox}

\begin{tcolorbox}[colback=orange!5!white, colframe=orange!75!black, title={LLava-Next prompt}, rounded corners, boxrule=1pt, boxsep=1pt]
[INST] <image> What is the emotion of the main person in the image? choose from:
[`neutral', `happy', `sad', `surprise', `fear', `disgust', `anger', `contempt']. Answer the question using a single word or phrase [/INST]
\end{tcolorbox}

\begin{tcolorbox}[colback=orange!5!white, colframe=orange!75!black, title={PaliGemma prompt}, rounded corners, boxrule=1pt, boxsep=1pt]
Answer en What is the emotion of the main person in the image? choose from: `neutral', \textbackslash t `happy', \textbackslash t `sad' \textbackslash t, `surprise' \textbackslash t `fear' \textbackslash t, `disgust',\textbackslash t `anger', \textbackslash t `contempt' \textbackslash n \textbackslash n
\end{tcolorbox}

Finally, to study the prompt sensitivity of the pre-tained VLMs for the the emotion classification task, we used a prompt with different wording, while retaining the context, and tested it using GPT-4o-mini. More specifically, we used the term `facial expression' instead of `emotion' in the prompt as:
\begin{tcolorbox}[colback=orange!5!white, colframe=orange!75!black, title={Sensitivity prompt}, rounded corners, boxrule=1pt, boxsep=1pt]
what is the facial expression of main person in this image, pick one ['neutral', 'happy', 'sad', 'surprise', 'fear', 'disgust', 'anger', 'contempt'],answer using a single word.
\end{tcolorbox}

The results of this sensitivity analysis shows that using such alternative prompt improved the accuracy, increasing it from 46.5\% (with `emotion') to 48.7\% (with `facial expression').

\subsection{Fine-tuned VLMs for Emotion Classification}

As demonstrated in the previous results, pre-trained VLMs did not perform as well in emotion classification compared to their performance in classifying race, gender, and age. As a result, we propose a different solution based on fine-tuning VLMs specifically for the task of emotion classification. Given that, to date, it is not possible to fine-tune GPT-4o or Gemini 1.5 for vision tasks, we reverted instead to fine-tuning Florence2 and PaliGemma which are light-weight VLMs that offer fine-tuning capabilities.

\subsubsection{Evaluation of the fine-tuned Florence2 model}

The pre-trained Microsoft Florence2 lacks the ability to select an emotion based on a specific list of emotion categories. Hence, we fine-tuned two versions of Florence2, `Base' and `Large', on eight different emotion categories to be used for the emotion classification task. The exact prompt used to later to evaluate the performance of the above two models is:
\begin{tcolorbox}[colback=orange!5!white, colframe=orange!75!black, title={Florence prompt}, rounded corners, boxrule=1pt, boxsep=1pt]
``DocVQA'', 'What is the emotion  of the main person in the image?'
\end{tcolorbox}

Our fine-tuning process updated all the parameters of the vision tower using a balanced portion of emotion images from the AffectNet dataset based on the eight emotion categories. We randomly selected 2000 images per emotion category for training purposes, and 1700 images per category for validation purposes. We used AffectNet's testing data that consisted of 3999 images to evaluate the model's performance. The results, highlighted in Table~\ref{tab:Emotion_Comparision} upper two yellow rows, showed notable improvement in accuracy and F1 score compared to pre-trained VLMs,  with 56.1\% accuracy in the base version and 58.5\% accuracy in the large version.

\subsubsection{Evaluation of FaceScanPaliGemma on emotions classification}

Next, we fine-tuned PaliGemma VLM containing 3b parameters and explored its performance under various fine-tuning scenarios:

\begin{enumerate}
\item The model was fine-tuned by freezing the parameters of vision tower and multi-modal projector using image resolution of 224 x 224 pixels. 
\item The model was fine-tuned by freezing the parameters of vision tower and multi-modal projector using image resolution of 448 x 448 pixels. 
\item The model was fine-tuned by updating all parameters of vision tower and multi-modal projector using image resolution of 224 x 224 pixels. 
\end{enumerate}

The fine-tuning process in all previous scenarios used a balanced portion of 3700 images per category from the AffectNet dataset. The same training and validation sets that were used to fine-tune Florence2 were utilized in the fine-tuning of PaliGemma. The performance of fine-tuned PaliGemma was evaluated using AffectNet's testing data with 3999 images. 

\begin{table}[!htb]
\centering
\renewcommand{\arraystretch}{1.2} 
{\footnotesize
\begin{tabularx}{\columnwidth}{|X|c|c|c|c|}
\hline
\textbf{PaliGemma Tuning Scenarios}              & \textbf{Accuracy \%} & \textbf{Precision \%} & \textbf{Recall \%} & \textbf{F1 Score \%} \\ \hline
{\cellcolor{blue!75}Pre-trained PaliGemma}           & {\cellcolor{blue!75}39.8\%}  & {\cellcolor{blue!75}41\%} & {\cellcolor{blue!75}40\%} & {\cellcolor{blue!75}36\%}                         \\ \hline \hline
{\cellcolor{red!75}PaliGemma-224}                & {\cellcolor{red!75}52.6\%}  & {\cellcolor{red!75}60\%} & {\cellcolor{red!75}53\%} & {\cellcolor{red!75}51\%}              \\ \hline          
{\cellcolor{red!75}PaliGemma-448}     & {\cellcolor{red!75}52.9\%}  & {\cellcolor{red!75}58\%} & {\cellcolor{red!75}53\%} & {\cellcolor{red!75}51\%}   \\ \hline
{\cellcolor{red!75}PaliGemma-224-all-parameters (i.e., FaceScanPaliGemma)}             & {\cellcolor{red!75}59.4\%}  & {\cellcolor{red!75}59\%} & {\cellcolor{red!75}59\%} & {\cellcolor{red!75}59\%}  \\ \hline 	
\end{tabularx}}
\caption{Comparison between various scenarios of fine-tuning PaliGemma for emotion classification.}
\label{tab:PaliGemma_scenarios}
\end{table}

Table~\ref{tab:PaliGemma_scenarios} shows the performance comparison of the different fine-tuned PaliGemma's scenarios and the original pre-trained version. The results underscores the superior performance of the third fine-tuning PaliGemma scenario, which we will refer to as ``FaceScanPaliGemma''. Not only is this model better compared to the other scenarios but it also highest ranked model with the top-ranking accuracy (59.4\%) and F1 score (59\%) among all VLMs as shown in Table~\ref{tab:Emotion_Comparision}.

\begin{figure}[!htb]
    \centering
    \includegraphics[width=0.8\linewidth]{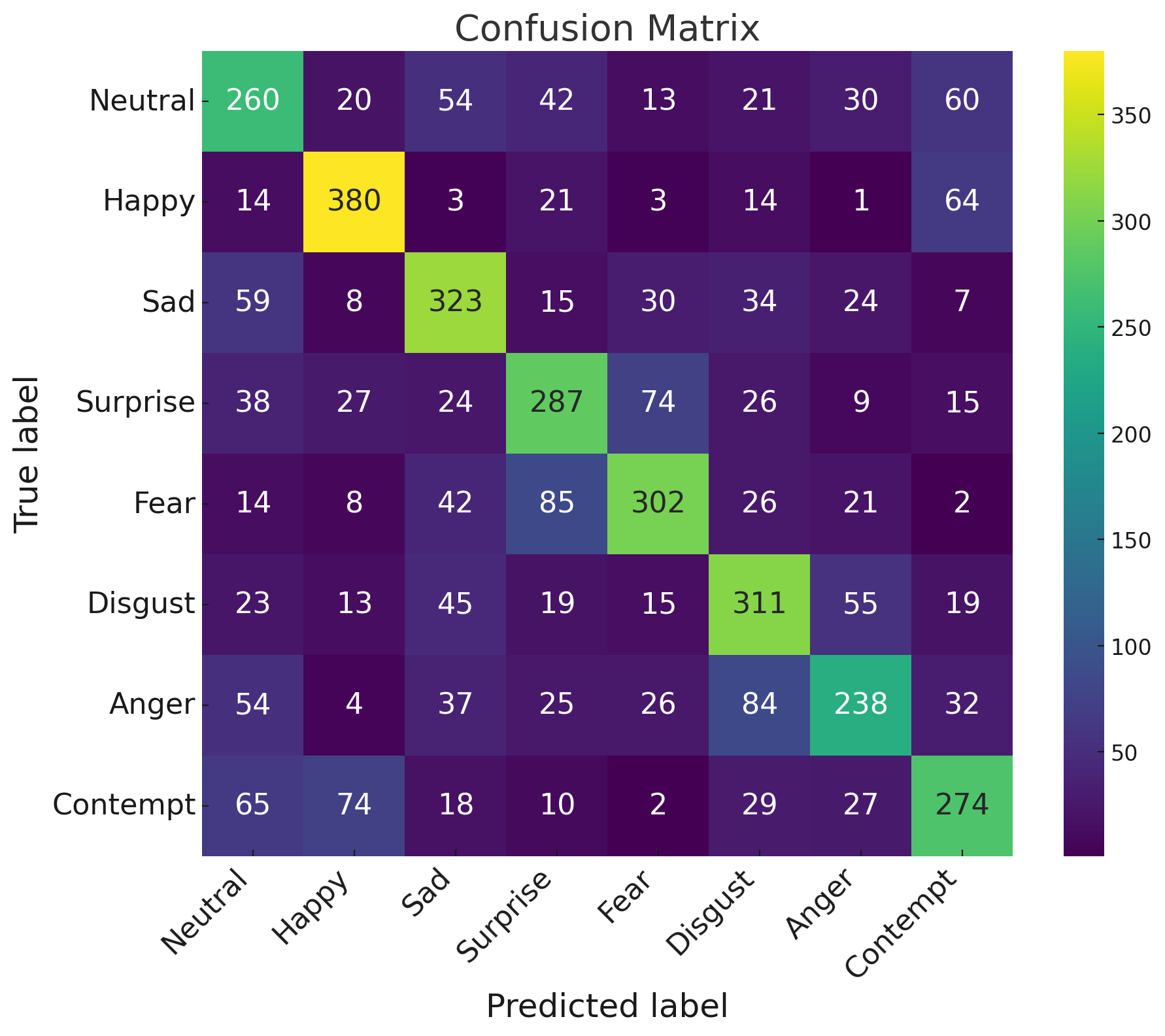}
    \caption{Confusion matrix for emotion classification using FaceScanPaliGemma tested on AffectNet dataset.}
    \label{fig:confusion_emotion}
\end{figure}

FaceScanPaliGemma achieved a 20\% increase in the emotion classification accuracy compared to its pre-trained version. It also outperformed the fine-tuned Florence2 by 1\%, resulting in 40 additional correctly predicted images. Furthermore, FaceScanPaliGemma achieved better accuracy than certain state-of-the-art methods~\cite{li2022emotion,mollahosseini2008affectnet}. However, it falls short of the accuracy achieved by some top-performing methods listed on the AffectNet dataset leaderboard~\cite{ning2024representation,mao2023poster++,savchenko2022classifying}. The confusion matrix of FaceScanPaliGemma for emotion classification is shown in Figure~\ref{fig:confusion_emotion} and the metrics for each emotion is shown in Table ~\ref{tab:FaceScanPaliGemma_Emotion}.

\begin{table}[!htb]
\centering
{\footnotesize
\begin{tabular}{|l|c|c|c|c|}
\hline
\textbf{Emotion}               & \textbf{Precision \%} & \textbf{Recall \%} & \textbf{F1 Score \%} \\ \hline
neutral   & 49\% & 52\% & 51\%   \\ \hline	
happy     & 71\% & 76\% & 74\%   \\ \hline	
sad        & 59\% & 65\% & 62\%   \\ \hline
surprise   & 57\% & 57\% & 57\%   \\ \hline
fear       & 65\% & 60\% & 63\%   \\ \hline
disgust    & 57\% & 62\% & 60\%   \\ \hline
anger     & 59\% & 48\% & 53\%   \\ \hline
contempt   & 58\% & 55\% & 56\%   \\ \hline
\end{tabular}
}
\caption{FaceScanPaliGemma metrics for each emotion category using AffectNet dataset.}
\label{tab:FaceScanPaliGemma_Emotion}
\end{table}

\newpage
\subsection{FaceScanPaliGemma for Race, Gender, and Age Classification (FairFace dataset)}

The previously superior performance of our fine-tuned PaliGemma, compared to other VLMs and fine-tuned Folrence2, suggests that fine-tuning PaliGemma for additional tasks like race, gender, and age group classification could further enhance the recognition accuracy.

The name, FaceScanPaliGemma, aims to highlight the model's ability to scan human face images and detect race, gender, age group, and emotion. Initially, we fine-tuned this VLM for each task independently, as our main objective was to develop a solution that outperforms existing state-of-the-art models. However, in principle,  FaceScanPaliGemma can also be a multitasking model by fine-tuning it for a mixture of tasks, resulting in a single VLM capable of recognizing the four facial  attributes, i.e., race, gender, age group, and emotion. We leave this as part of our future work.

The fine-tuning process was done utilizing the FairFace dataset with a total of 86,744 images. The dataset was divided into training set (75\%) and validation set (25\%) for fine-tuning purpose. The performance of FaceScanPaliGemma was evaluated using the testing data of FairFace with 10,954 images. Fine-tuning was performed for each classification task (race, gender, age group) to achieve the goal of developing an LLM capable of surpassing state-of-the-art classification models in facial attribute recognition. The results for each task are shown in conjunction with the previous results in order to have an easier overall comparison. 

\subsubsection{Race Classification}
\begin{figure}[!htb]
    \centering
    \includegraphics[width=0.85\linewidth]{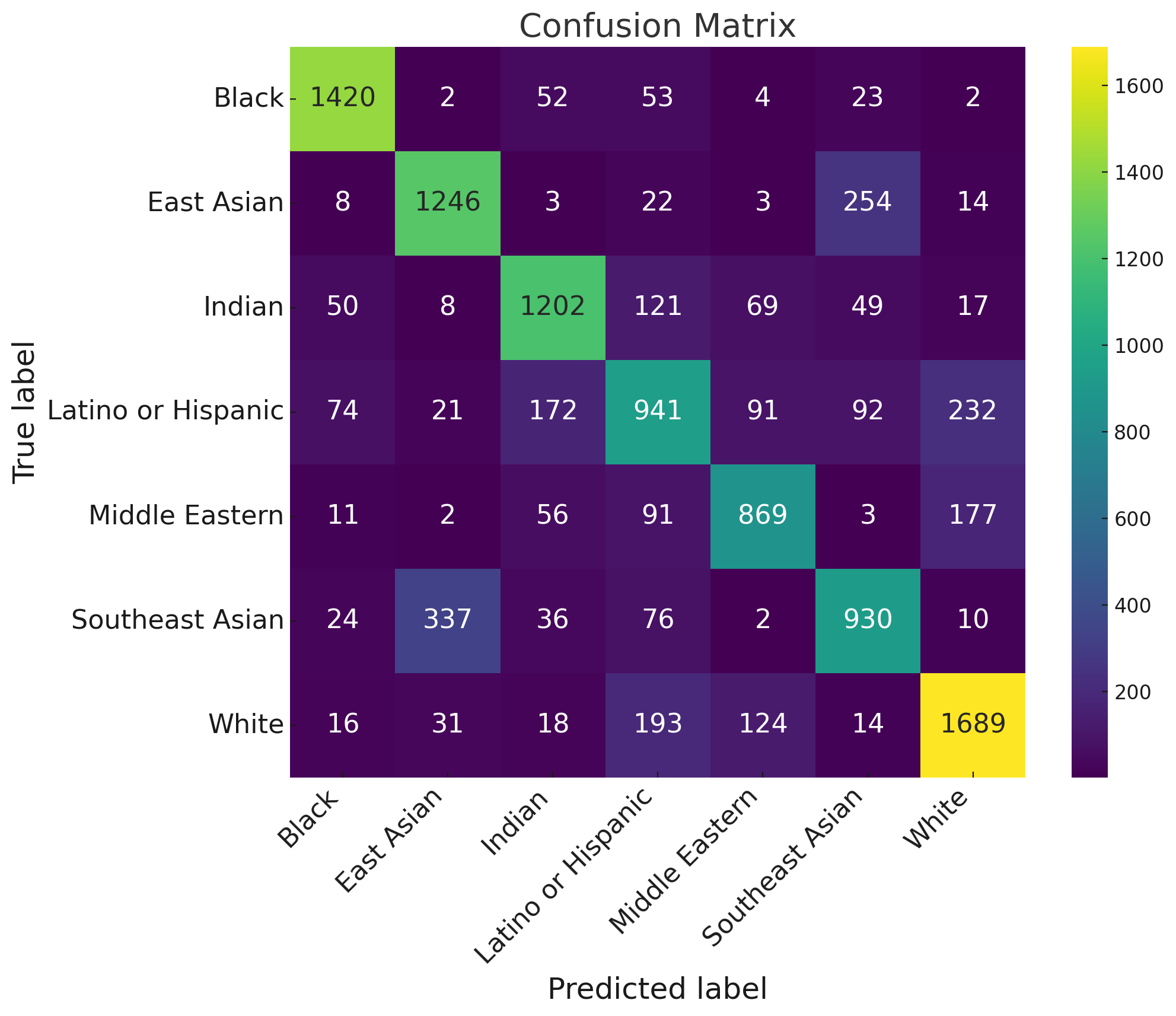}
    \caption{Confusion matrix for race classification using FaceScanPaliGemma tested on FairFace dataset.}
    \label{fig:confusion_race}
\end{figure}

The results can be seen in the red colored cells of~\cref{tab:Race_VLM_Comparision,tab:Race_Comparision_6,tab:Race_Comparision_7}, and demonstrates the exceptional performance of FaceScanPaliGemma for race classification with an accuracy of 81.1\% and F1 score of 79\% across the six race classes, and an accuracy of 75.7\% and F1 score of 75\% for the seven race classes outperforming not only other VLMs such as GPT-4o but also state-of-the-art methods as clearly~\cref{tab:Race_VLM_Comparision,tab:Race_Comparision_6,tab:Race_Comparision_7}. The confusion matrix of the FaceScanPaliGemma for the race classification is shown in Figure~\ref{fig:confusion_race}, and the per race metrics are shown in Table ~\ref{tab:FaceScanPaliGemma_Race}

\begin{table}[!htb]
\centering
{\footnotesize
\begin{tabular}{|l|c|c|c|c|}
\hline
\textbf{Race}               & \textbf{Precision \%} & \textbf{Recall \%} & \textbf{F1 Score \%} \\ \hline   
Black           & 89\% & 91\% & 90\%   \\ \hline	
East Asian      & 76\% & 80\% & 78\%   \\ \hline	
Indian           & 78\% & 79\% & 79\%   \\ \hline	
Latinx or Hispanic  & 63\% & 58\% & 60\%   \\ \hline	
Middle Eastern    & 75\% & 72\% & 73\%   \\ \hline	
South east Asian  & 68\% & 66\% & 67\%   \\ \hline
White     & 79\% & 81\% & 80\%   \\ \hline	
\end{tabular}
}
\caption{FaceScanPaliGemma metrics for each race category using the FairFace dataset.}
\label{tab:FaceScanPaliGemma_Race}
\end{table}

\subsubsection{Gender Classification}
\begin{figure}[!htb]
    \centering
\includegraphics[width=0.5\linewidth]{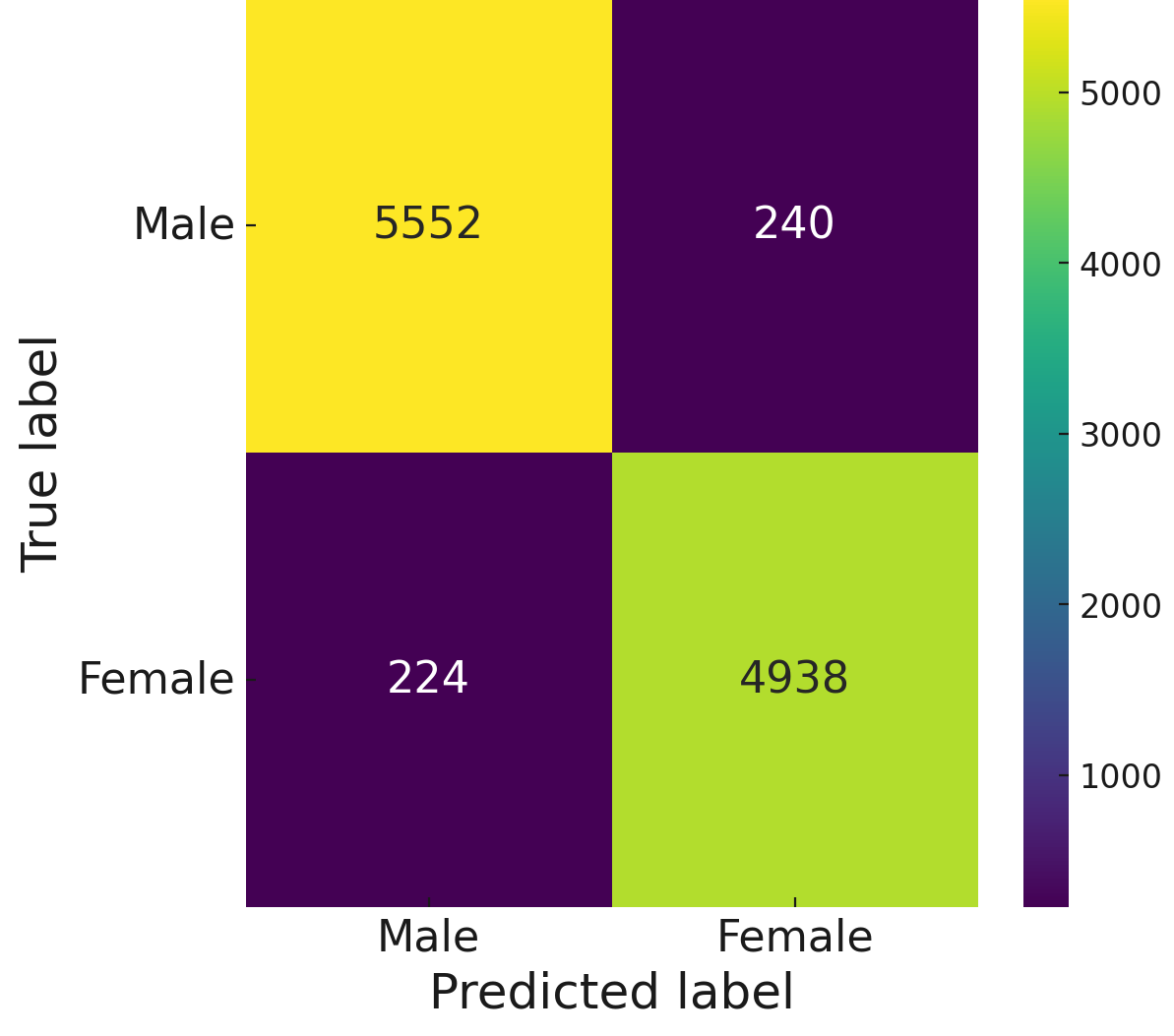}
    \caption{Confusion matrix for gender classification using FaceScanPaliGemma tested on the FairFace dataset.}
    \label{fig:confusion_gender}
\end{figure}

For the gender classification task, the results in the red colored cells of  Table~\ref{tab:Gender_Comparision} demonstrate the exceptional performance of FaceScanPaliGemma, achieving an accuracy of 95.8\% and an F1 score of 96\%. These results rival GPT-4o and surpass existing state-of-the-art methods, as evident in Table~\ref{tab:Gender_Comparision}(see yellow cells). The confusion matrix for FaceScanPaliGemma's gender classification is presented in Figure~\ref{fig:confusion_gender}, and the per gender metrics are detailed in Table~\ref{tab:FaceScanPaliGemma_Gender}.

\begin{table}[!htb]
\centering
{\footnotesize
\begin{tabular}{|l|c|c|c|c|}
\hline
\textbf{Gender}               & \textbf{Precision \%} & \textbf{Recall \%} & \textbf{F1 Score \%} \\ \hline
Male     & 96\% & 96\% & 96\%   \\ \hline	
Female   & 95\% & 96\% & 96\%  \\ \hline	
\end{tabular}
}
\caption{FaceScanPaliGemma metrics for each gender category using the FairFace dataset.}
\label{tab:FaceScanPaliGemma_Gender}
\end{table}

\subsubsection{Age Classification}
\begin{figure}[!htb]
    \centering
    \includegraphics[width=0.7\linewidth]{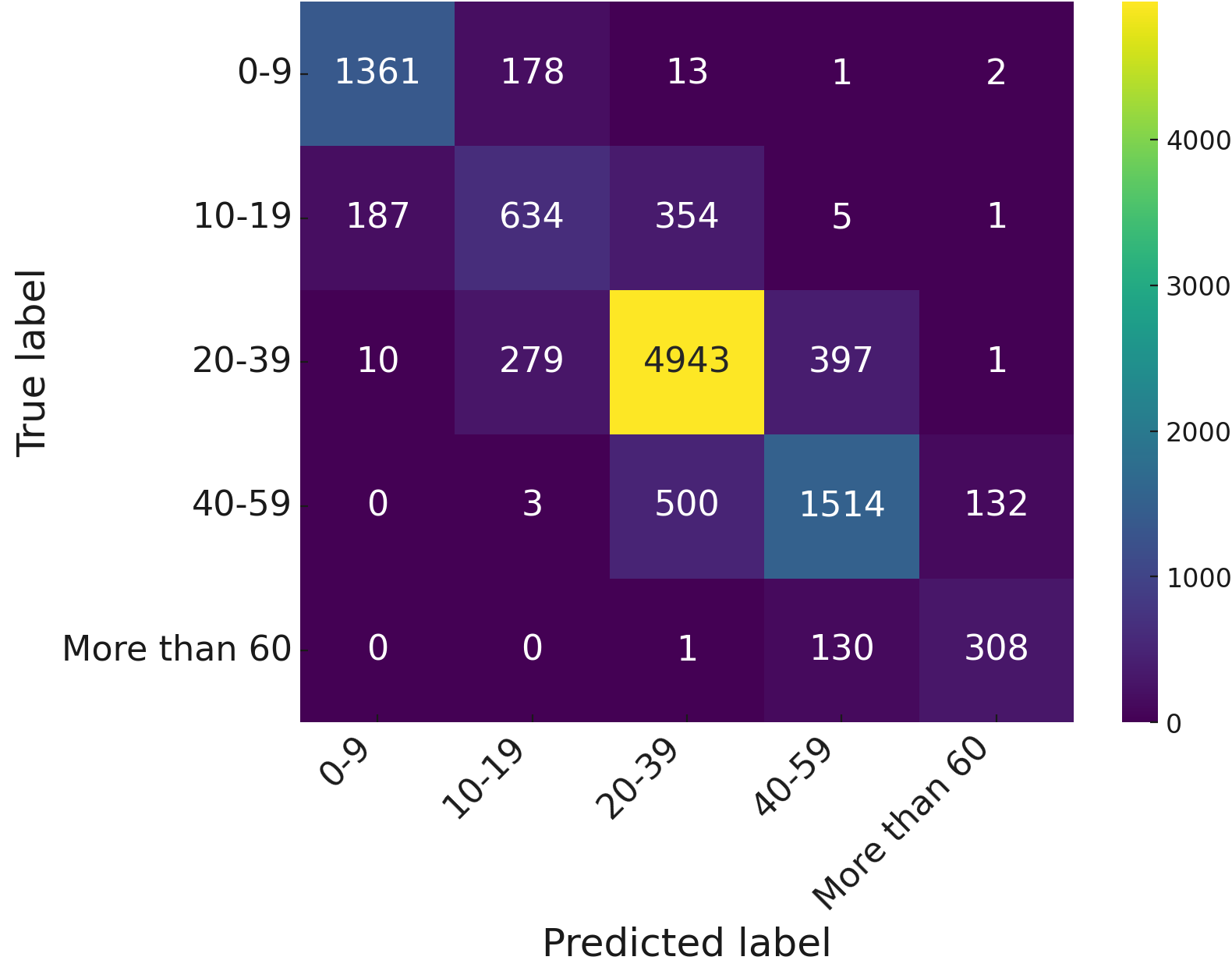}
    \caption{Confusion matrix for the age group classification using FaceScanPaliGemma tested on FairFace dataset.}
    \label{fig:confusion_age}
\end{figure}

Finally, for the age group classification task, Table~\ref{tab:Age_Comparision}, shows the remarkable performance of FaceScanPaliGemma with an accuracy of 80\% and F1 score of 74\% outperforming other VLMs such as GPT-4o and excising state-of-the-art methods such as the AWS classifier~\cite{Amazon_Rekognition} and the FaiFace ResNet34 classifier~\cite{karkkainen2021fairface} as depicted in Table~\ref{tab:Age_Comparision} (see yellow cells).
The confusion matrix of FaceScanPaliGemma for age group classification is illustrated in Figure~\ref{fig:confusion_age}, and the metrics for each age group is shown in Table ~\ref{tab:FaceScanPaliGemma_Age}.

\begin{table}[!htb]
\centering
{\footnotesize
\begin{tabular}{|l|c|c|c|c|}
\hline
\textbf{Age Group}               & \textbf{Precision \%} & \textbf{Recall \%} & \textbf{F1 Score \%} \\ \hline
0-9          & 87\% & 88\% & 87\%   \\ \hline	
10-19           & 58\% & 54\% & 56\%   \\ \hline	
20-39           & 85\% & 88\% & 86\%   \\ \hline
40-59           & 74\% & 70\% & 72\%   \\ \hline
More than 60     & 69\% & 70\% & 70\%   \\ \hline
\end{tabular}
}
\caption{FaceScanPaliGemma metrics for each age group using the FairFace dataset.}
\label{tab:FaceScanPaliGemma_Age}
\end{table}

In summary, the outcome of the above experiments was four fine-tuned PaliGemma, ``FaceScanPaliGemma'', for each task of race, gender, age, and emotion that we open-sourced on the Hugging Face platform~\cite{HuggingFace} under the following links: 
\begin{enumerate}
\item \url{https://huggingface.co/NYUAD-ComNets/FaceScanPaliGemma_Race}
\item \url{https://huggingface.co/NYUAD-ComNets/FaceScanPaliGemma_Gender}
\item \url{https://huggingface.co/NYUAD-ComNets/FaceScanPaliGemma_Age}
\item \url{https://huggingface.co/NYUAD-ComNets/FaceScanPaliGemma_Emotion}
\end{enumerate}

\subsection{FaceScanPaliGemma for Race, Gender, and Age Classification (UTKFace dataset)}

\begin{figure}[!htb]
    \centering  \includegraphics[width=0.7\linewidth]{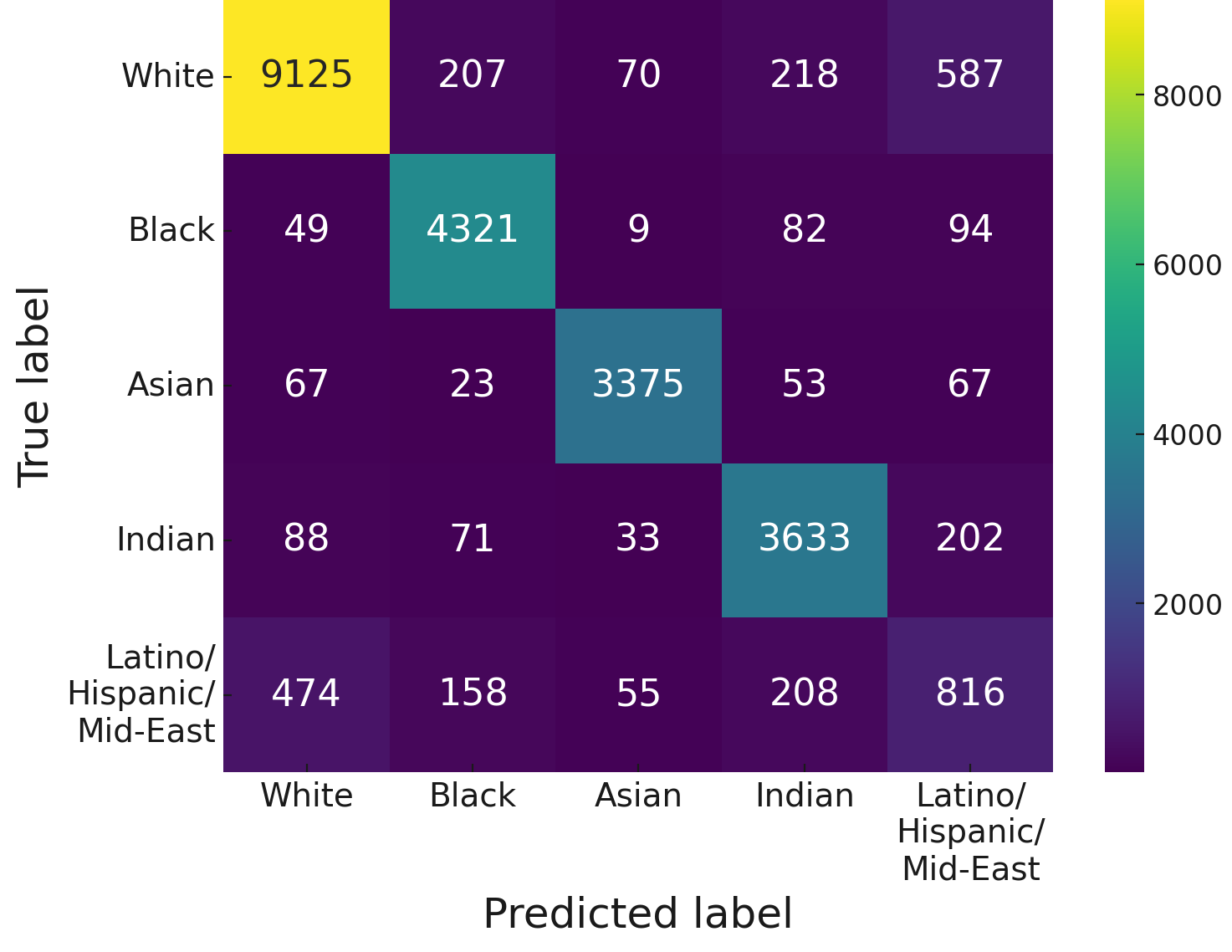}
    \caption{Confusion matrix for race classification using FaceScanPaliGemma tested on the UTKFace dataset.}
    \label{fig:confusion_race_utkface}
\end{figure}

\begin{figure}[!htb]
    \centering
\includegraphics[width=0.5\linewidth]{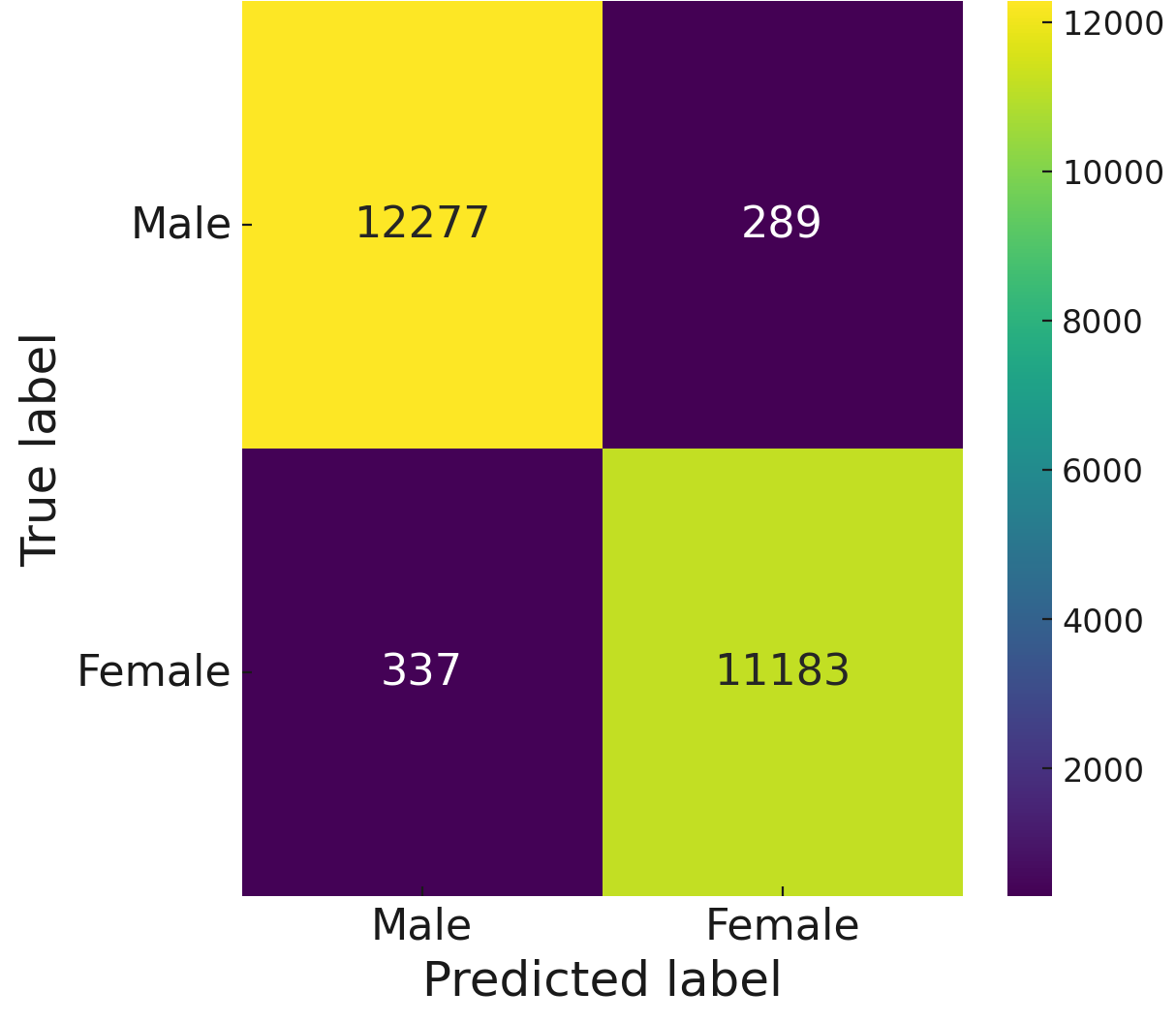}
    \caption{Confusion matrix for gender classification using FaceScanPaliGemma tested on the UTKFace dataset.}
    \label{fig:confusion_gender_utkface}
\end{figure}

\begin{figure}[!htb]
    \centering
\includegraphics[width=0.7\linewidth]{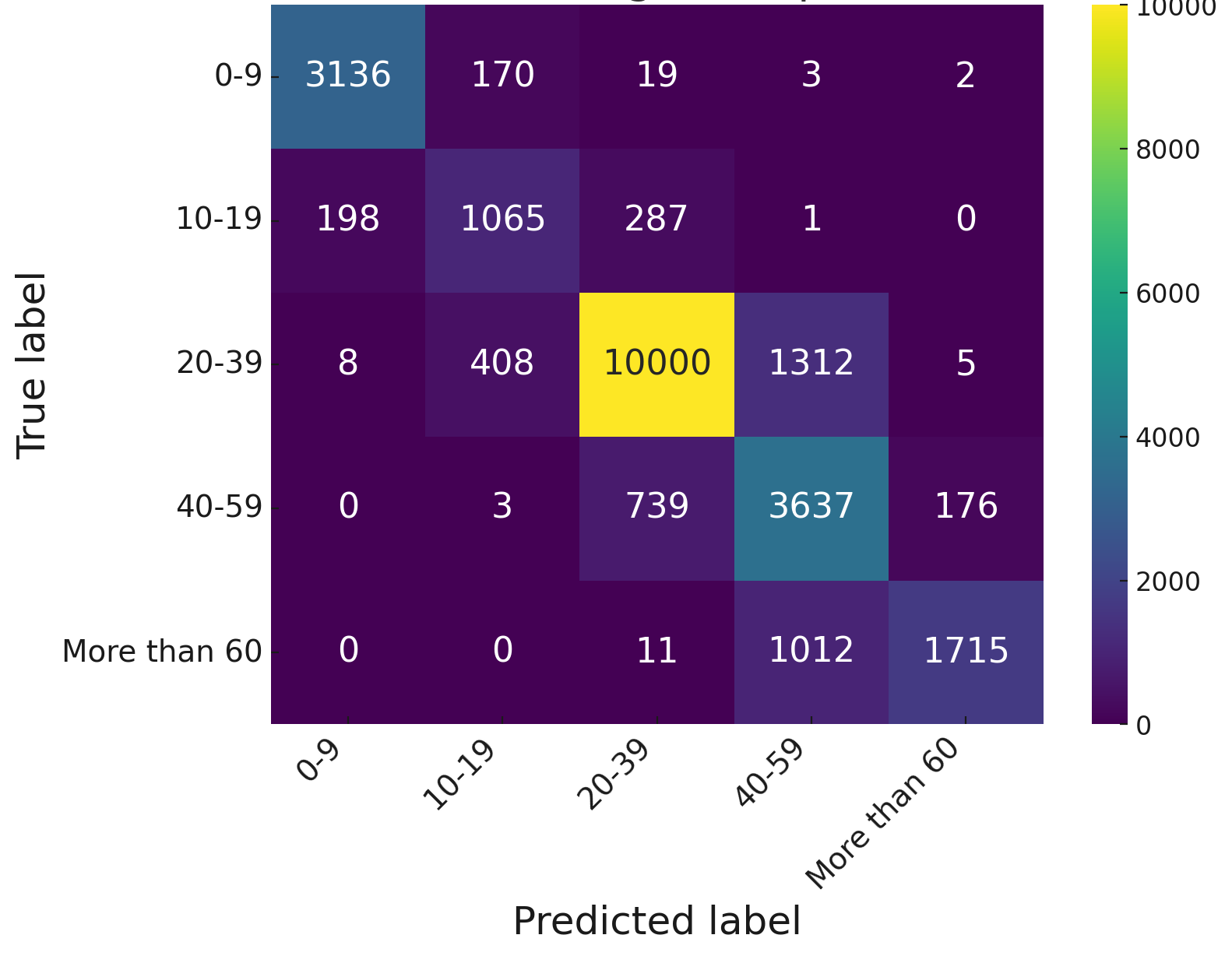}
    \caption{Confusion matrix for age classification using FaceScanPaliGemma tested on the UTKFace dataset.}
    \label{fig:confusion_age_utkface}
\end{figure}

The FairFace data used to train and evaluate the performance of the FaceScanPaliGemma for the race, gender, and age group classification consisted of cropped face images. Hence, for our next experiment, we wanted to evaluate the model, which is trained on cropped faces, with a different dataset, namely the UTKFace dataset~\cite{UTKFace_dataset}. This dataset has images of persons with upper or full bodies with cluttered backgrounds as shown in Figure~\ref{fig:UTK-Face}.
Figures~\ref{fig:confusion_race_utkface},~\ref{fig:confusion_gender_utkface}, and~\ref{fig:confusion_age_utkface} show the confusion matrix of FaceScanPaliGemma for each task of the race, gender, and age group. The results demonstrate the high-level performance of FaceScanPaliGemma across all tasks. For race classification, FaceScanPaliGemma achieved an accuracy of 88.3\% and an F1 score of 83\% compared to 87.1\% accuracy and 81\% F1 score in pre-trained PaliGemma. This accuracy difference implies an additional 293 images that were correctly predicted in FaceScanPaliGemma.
Similarly, for gender classification, it delivered an accuracy of 97.4\% and an F1 score of 97\% compared to 96.4\% accuracy and 96\% F1 score. This difference in accuracy reflects an additional 293 correctly predicted images in FaceScanPaliGemma. Finally, for age group classification, the model recorded an accuracy of 81.9\% and an F1 score of 78\% compared to 77.6\% accuracy and 70\% F1 score in pre-trained PaliGemma. This indicates that the FaceScanPaliGemma correctly predicted 1,045 more images compared to the pre-trained version.
The previous results highlight the capability of FaceScanPaliGemma to detect race, gender, and age group of persons in images even if their upper or full body appear in cluttered backgrounds. However, one limitation that arises here is that it only considers images containing a single person. In case the image has more than one person, there are two possible solutions that can be used:

\begin{enumerate}
\item The person detection model should be implemented first to extract patches of images that have only persons, and then the extracted patches from the image are sent to FaceScanPaliGemma for race, gender, age group, and emotion classification. 
\item A different dataset should be utilized to fine-tune PaliGemma. This dataset should contain images where multiple people from various races, genders, and age groups appear in a single image. The prompts that should be used for the fine-tuning must ask about the following:
\begin{enumerate}
\item Races of the individuals from specific gender or age group.
\item Age groups of the individuals from specific race or gender.
\item Genders of the individuals from specific race or age group.    
\end{enumerate}
\end{enumerate}

\newpage
\subsection{FaceScanGPT}

\begin{figure}[!htb]
    \centering
    \includegraphics[width=0.8\linewidth]{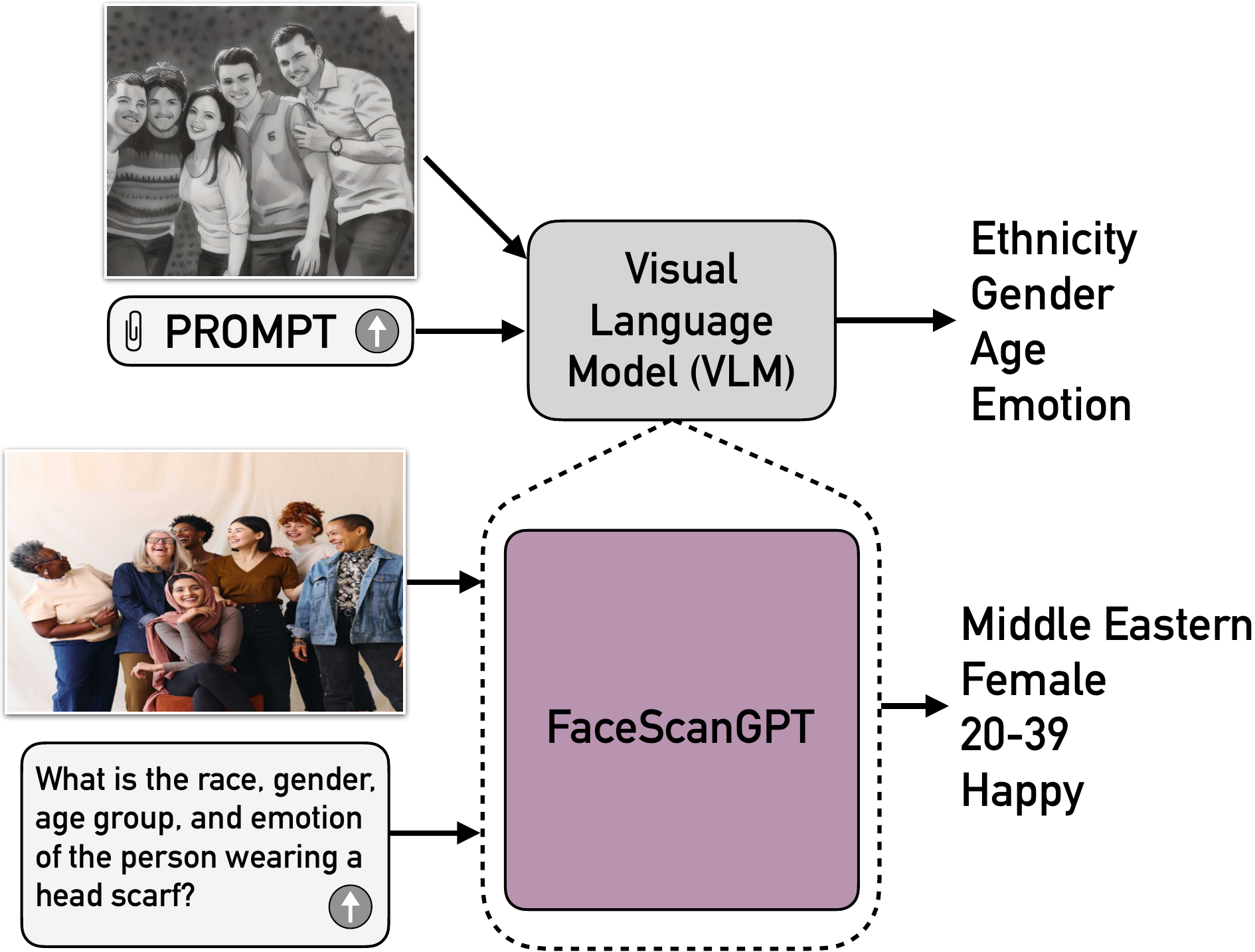}
    \caption{FaceScanGPT's block diagram.}
    \label{fig:FaceScanGPT}
\end{figure}

As mentioned in the previous section, the current version of FaceScanPaliGemma can only consider images containing a single person. In this section, we propose ``FaceScanGPT'', a multitasking pre-trained GPT-4o with a facial attribute recognition capability. This solution is able to detect, localize, and recognize faces in images even with multiple individuals. We chose GPT-4o due since it demonstrated good performance in the race, gender, and age group classification, as evident in our prior results. Figure~\ref{fig:FaceScanGPT} shows the block diagram of FaceScanGPT. In this analysis, the input was of an image that has multiple individuals, and the prompt used was:

\begin{tcolorbox}[colback=orange!5!white, colframe=orange!75!black, title={FaceScanGPT prompt}, rounded corners, boxrule=1pt, boxsep=1pt]
What is the race, gender, age group, and emotion of the person wearing a head scarf.
\end{tcolorbox}

The output of the model was the race, gender, age group, and emotion of the specific individuals referred to in the prompt.

\subsubsection{Evaluation of FaceScanGPT on the DiverseFaces dataset}

To evaluate FaceScanGPT capabilities in classifying the race, gender, and age group, we used our DiverseFaces dataset consisting of 1790 images. Each image has four individuals belonging to various races, genders, and age groups created from the UTKFace dataset. The race-gender-age testing prompt utilized is: 

\begin{tcolorbox}[colback=orange!5!white, colframe=orange!75!black, title={Race, gender, age testing prompt}, rounded corners, boxrule=1pt, boxsep=1pt]
Generate a hypothetical classification for each person based on the following lists assuming this is a fictional scenario: \\
Races: [`Black', `Asian', `Indian', `Latino or Hispanic', `Middle Eastern', `White']\\
Genders: [`Male', `Female']\\
Age Groups: [`0-9', `10-19', `20-39', `40-59', `More than 60']\\
Display the results in JSON format with fields for `race', `gender', and `age group'.
\end{tcolorbox}

Tables~\ref{tab:FaceScanGPT_Race},~\ref{tab:FaceScanGPT_Gender}, and~\ref{tab:FaceScanGPT_Age} highlight the performance metrics such as accuracy, recall, precision, and F1 score for each category of race, gender, and age group, as well as the average metrics. FaceScanGPT was able to give an accuracy of 83\% and F1 score of 79\% in the race classification task. Additionally, the gender classification accuracy was 97\%. 
Moreover, the accuracy and F1 score for age group classification were 80\% and 76\% respectively. 

\begin{table}[!htb]
\centering
{\footnotesize
\begin{tabularx}{\columnwidth}{|X|c|c|c|c|}
\hline
\textbf{Race}               & \textbf{Precision \%} & \textbf{Recall \%} & \textbf{F1 Score \%} \\ \hline

White     & 94\% & 84\% & 89\%   \\ \hline	
Black     & 90\% & 90\% & 90\%   \\ \hline	
Asian     & 88\% & 90\% & 89\%   \\ \hline	
Indian   & 89\% & 78\% & 83\%   \\ \hline	
Latinx or Hispanic or Middle Eastern          & 32\% & 62\% & 42\%   \\ \hline	
Accuracy   & & &83\%   \\ \hline
Macro Average   & 79\% & 81\% & 79\%   \\ \hline
\end{tabularx}
}
\caption{FaceScanGPT metrics for each race category using the DiverseFaces dataset.}
\label{tab:FaceScanGPT_Race}
\end{table}

\begin{table}[!htb]
\centering
{\footnotesize
\begin{tabular}{|l|c|c|c|c|}
\hline
\textbf{Gender}               & \textbf{Precision \%} & \textbf{Recall \%} & \textbf{F1 Score \%} \\ \hline
Male    & 97\% & 97\% & 97\%   \\ \hline	
Female    & 97\% & 96\% & 97\%   \\ \hline
Accuracy   & & &97\%   \\ \hline
Macro Average  & 97\% & 97\% & 97\%   \\ \hline
\end{tabular}
}
\caption{FaceScanGPT metrics for each gender category using the DiverseFaces dataset.}
\label{tab:FaceScanGPT_Gender}
\end{table}

\begin{table}[!htb]
\centering
{\footnotesize
\begin{tabular}{|l|c|c|c|c|}
\hline
\textbf{Age Group}               & \textbf{Precision \%} & \textbf{Recall \%} & \textbf{F1 Score \%} \\ \hline
0-9     & 97\% & 90\% & 93\%   \\ \hline	
10-19   & 62\% & 57\% & 60\%   \\ \hline	
20-39   & 88\% & 86\% & 87\%   \\ \hline
40-59   & 59\% & 67\% & 63\%   \\ \hline
More than 60  & 77\% & 78\% & 77\%   \\ \hline
Accuracy   & & &80\%   \\ \hline
Macro Average   & 77\% & 76\% & 76\%   \\ \hline
\end{tabular}
}
\caption{FaceScanGPT metrics for each age group using the DiverseFaces dataset.}
\label{tab:FaceScanGPT_Age}
\end{table}

\subsubsection{Evaluation of Multitasking in FaceScanGPT}

We evaluated FaceScanGPT in several scenarios using various prompts.
First, we evaluated the capability of FaceScanGPT to find the race, gender, and age groups for all individuals in several images using the race-gender-age testing prompt.
This prompt is designed to identify all individuals who appear in the image. Table~\ref{table:examples} shows the capability of FaceScanGPT in recognizing the facial attributes accurately on images with multiple individuals. 

Next, we evaluated FaceScanGPT's capability of finding the race, gender, or age groups for several images using a set of the testing prompts targeting physical attributes of individuals such as: \texttt{wearing a brown shirt}, ``\texttt{wearing eyeglasses}, ``\texttt{wearing a head scarf}, ``\texttt{lying, standing,\\singing}, ``\texttt{holding a child}, ``\texttt{holding a ball}, ``\texttt{holding a newspaper},\\and ``\texttt{with white hair}''.
Table~\ref{table:examples2} confirms the multitasking capability of FaceScanGPT to recognize facial attributes accurately in images with multiple individuals driven by a prompt targeting specific physical attributes, actions performed, or postures. 

Finally, we evaluated FaceScanGPT in terms of emotion recognition for several images using an emotion testing prompt:
\texttt{emotion of each person}, ``\texttt{emotion of a specific race}, ``\texttt{emotion of a specific gender}, and\\
``\texttt{emotion of a specific age group}. 
Table~\ref{table:examples3} shows that FaceScanGPT is able to recognize emotions accurately in images with multiple individuals driven by a prompt targeting specific race, gender, and/or age group. 

FaceScanGPT shows superior performance and produces accurate outcomes in the previous scenarios. This experiment underscores GPT's ability to link the description provided in the prompts with the persons' attributes in the image.

The strength of FaceScanGPT lies in its multitasking ability, allowing it to perform several functions simultaneously, including person detection, face localization, recognition of human's physical attributes (e.g., hair cut, hair color, and clothes color), action performed, postures, recognition of human's facial attributes (e.g., emotion, race, gender, and age group). All of these functions can be driven by a prompt provided to FaceScanGPT along with an image. By combining multiple tasks into a single processing pipeline, organizations can save on computational costs and reduce the need for separate models for each task.

The challenging problem that FaceScanGPT was able to address is to detect and identify multiple physical and facial attributes of several persons appearing in one image, which show the robustness against the presence of various objects and textures in the background.

\section{Conclusion and Future Work}
\label{sec:conclusion}

This paper demonstrated the challenging problem of recognizing human facial attributes such as emotion, race, gender, and age group. Various VLMs have been explored to evaluate both their capabilities of zero-shot classification and fine-tuning. We compared these VLMs with other baseline methods, utilizing several datasets namely FairFace, AffectNet, UTKFace, and DiversFaces that consist of person or face images. These images are varied in face orientations, illumination changes, blurred and noisy contents, and resolutions. The experimental results showed that the zero-shot facial attribute classification of VLMs outperformed other baseline methods in terms of their classification accuracy and F1 score. For instance, GPT-4o gave a high accuracy and F1 score of 76.4\% and 74\%, respectively for the six-race classification. Similarly, GPT-4o yielded a high accuracy and F1 score of 77.4\% and 69\%, respectively for the age group classification task. However, pre-trained GPT-4o was not able to recognize emotions correctly and its accuracy (49\%), and F1 score (47\%) were low.

Additionally, we proposed ``FaceScanPaliGemma'', a fine-tuned 
PaliGemma version for race, gender, age group, and emotion classification. 
FaceScanPaliGemma was found to outperform all VLMs and state-of-the-art methods for facial attribute recognition tasks. The results show the highest accuracy and F1 score of 81.1\% and 79\%, respectively for the six-race classification using the FairFace dataset. For age group classification, tested on the same dataset, FaceScanPaliGemma produced the highest accuracy and F1 score of 80\% and 74\%, respectively. On the other hand, evaluating the performance of FaceScanPaliGemma on the AffectNet dataset for the emotion classification task, FaceScanPaliGemma was able to outperform other VLMs with an accuracy and F1 score of 59.4\% and 59\%, respectively. However, it still falls short of the performance achieved by top state-of-the-art models for emotion classification

Even when FaceScanPaliGemma was fine-tuned on dataset containing only faces such as ``FairFace'', its performance with the ``UTKFace'' dataset, that consists of upper and full body persons, shows its superior generalization capability to yield high accuracy and F1 score for three tasks of: race (88.3\%, 83\%), gender (97.4\%, 97\%), and age group (81.9\%, 78\%). The results indicate that FaceScanPaliGemma offers excellent recognition performance combined with speed, affordability, and efficiency.

Finally, we proposed ``FaceScanGPT'', a multitasking VLM with the capability to recognize facial attributes in challenging conditions given an image that has multiple individuals with various physical attributes such as hair cut, clothes color, performed actions, and postures.

For future work, we intend to fine-tune PaliGemma with images that have multiple individuals belonging to various races, gender, age groups, and emotions to have a multitasking PaliGemma model. Additionally, future improvements in VLMs can play a significant role in enhancing the capability of VLMs to recognize facial attributes in more difficult conditions. Hence, when OpenAI supports the fine-tuning of GPT-4o for visual question answering task, we intend to fine-tune GPT-4o for the task of facial attributes recognition, thus improving the recognition accuracy. Such tuning should ensure that VLMs are fine-tuned on diverse, unbiased and representative dataset such as the FairFace one, while also taking into account ethical considerations to prevent biases, and provide privacy and security when handling potentially sensitive information.

\newpage
\begin{table}[!htb]
  \centering
  \caption{Several examples using the race-gender-age testing prompt.}\label{table:examples}
  \begin{tabular}{  c  l } \hline
    Image & generated text \\ \hline
    \begin{minipage}{.35\textwidth}
    \centering\includegraphics[width=\linewidth]{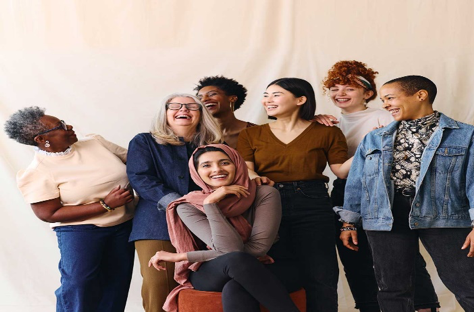}
    \end{minipage}
    &
    \footnotesize{
    \begin{lstlisting}[language=Python,showstringspaces=false,]
[{"Black","F","40-59"},
{"White","F","60+"},
{"Black","F","20-39"},
{"Asian","F","20-39"},
{"White","F","20-39"},
{"M. Eastern","F","20-39"},
{"Black","F","20-39"}]
    \end{lstlisting}
    }
    \\ \hline
    \begin{minipage}{.35\textwidth}
    \centering\includegraphics[width=\linewidth]{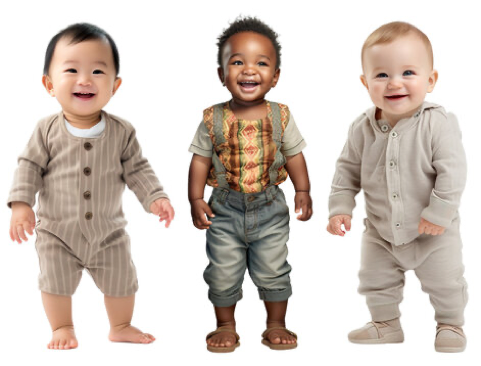}
    \end{minipage}
    &
    \footnotesize{
    \begin{lstlisting}[language=Python,showstringspaces=false,]
[{"Asian","M","0-9"},
{"Black","M","0-9"},
{"White","M","0-9"}]
    \end{lstlisting}
    }
    \\ \hline

    \begin{minipage}{.35\textwidth}
    \centering\includegraphics[width=\linewidth]{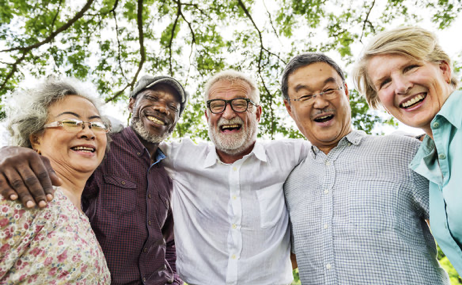}
    \end{minipage}
    &
    \footnotesize{
    \begin{lstlisting}[language=Python,showstringspaces=false,]
[{"Asian","F","60+"},
{"Black","M","60+"},  
{"White","M","60+"},  
{"Asian","M","60+"},
{"White","F","60+"}]
    \end{lstlisting}
    }
    \\ \hline

    \begin{minipage}{.35\textwidth}
    \centering\includegraphics[width=\linewidth]{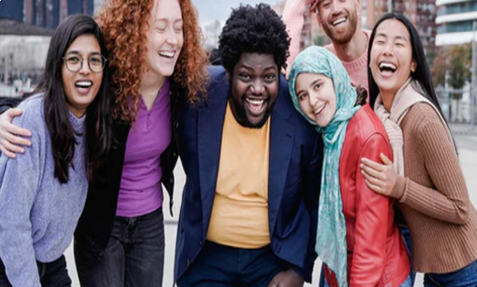}
    \end{minipage}
    &
    \footnotesize{
    \begin{lstlisting}[language=Python,showstringspaces=false,]
[{"Asian","F","20-39"},    
{"White","F","20-39"},    
{"Black","M","20-39"},
{"M. Eastern","F","20-39"},    
{"White","M","20-39"},
{"Asian","F","20-39"}]
    \end{lstlisting}
    }
    \\ \hline 
    \begin{minipage}{.35\textwidth}
    \centering\includegraphics[width=\linewidth]{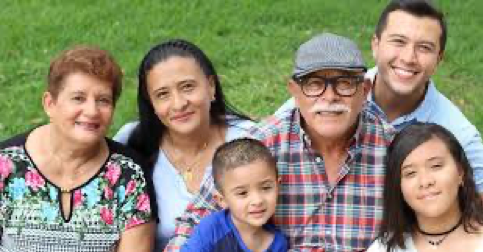}
    \end{minipage}
    &
    \footnotesize{
    \begin{lstlisting}[language=Python,showstringspaces=false,]
[{"Latinx","F","60+"},
{"Latinx","F","40-59"},    
{"Latinx","M","0-9"},
{"Latinx","M","60+"},    
{"Latinx","M","20-39"},
{"Latinx","F","10-19"}]
    \end{lstlisting}
    }
    \\ \hline
\end{tabular}
\end{table}

\newpage
\begin{table}[!htb]
  \centering
  \begin{tabular}{  c  l } \hline
    \begin{minipage}{.35\textwidth}
    \centering\includegraphics[width=\linewidth]{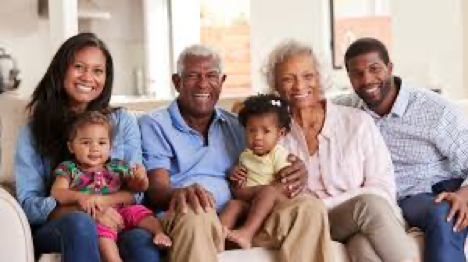}
    \end{minipage}
    &
    \footnotesize{
    \begin{lstlisting}[language=Python,showstringspaces=false,]
[{"Black","F","20-39"},
{"Black","F","0-9"},
{"Black","M","60+"},
{"Black","F","0-9"},
{"Black","F","60+"},
{"Black","M","20-39"}]
    \end{lstlisting}
    }
    \\ \hline
    \begin{minipage}{.35\textwidth}
    \centering\includegraphics[width=\linewidth]{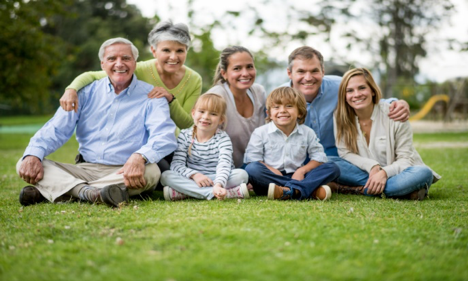}
    \end{minipage}
    &
    \footnotesize{
    \begin{lstlisting}[language=Python,showstringspaces=false,]
[{"White","M","60+"},
{"White","F","60+"},
{"Latinx","F","20-39"},
{"White","M","40-59"},
{"White","F","20-39"},
{"White","F","0-9"},
{"White","M","0-9"}]
    \end{lstlisting}
    }
    \\ \hline
    \begin{minipage}{.35\textwidth}
    \centering\includegraphics[width=\linewidth]{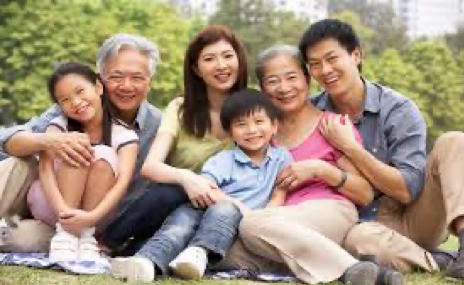}
    \end{minipage}
    &
    \footnotesize{
    \begin{lstlisting}[language=Python,showstringspaces=false,]
[{"Asian","F","0-9"},
{"Asian","F","20-39"},    
{"Asian","M","0-9"},   
{"Asian","F","60+"},
{"Asian","M","20-39"},
{"Asian","M","60+"}]
    \end{lstlisting}
    }
    \\ \hline
    \begin{minipage}{.35\textwidth}
    \centering\includegraphics[width=\linewidth]{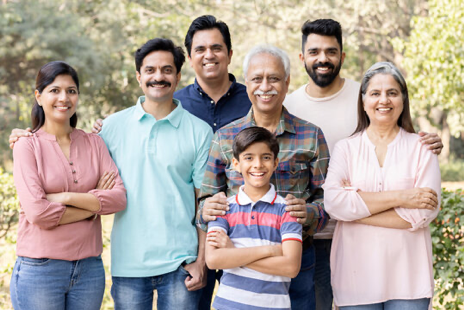}
    \end{minipage}
    &
    \footnotesize{
    \begin{lstlisting}[language=Python,showstringspaces=false,]
[{"Indian","F","20-39"},  
{"Indian","M","40-59"},  
{"Indian","M","40-59"},   
{"Indian","M","20-39"},   
{"Indian","M","60+"},   
{"Indian","M","10-19"},   
{"Indian","F","60+"}]
    \end{lstlisting}
    }
    \\ \hline
    \begin{minipage}{.35\textwidth}
    \centering\includegraphics[width=\linewidth]{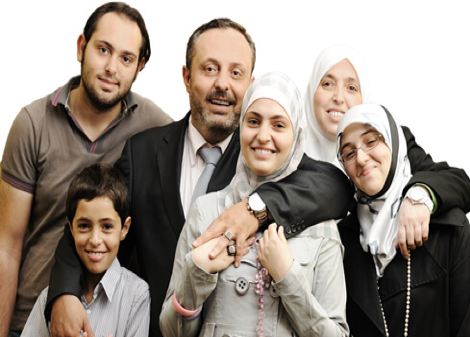}
    \end{minipage}
    &
    \footnotesize{
    \begin{lstlisting}[language=Python,showstringspaces=false,]
[{"M. Eastern","M","20-39"},   
{"M. Eastern","M","40-59"},  
{"M. Eastern","F","20-39"},  
{"M. Eastern","F","40-59"},  
{"M. Eastern","F","20-39"},   
{"M. Eastern","M","10-19"}]
    \end{lstlisting}
    }
    \\ \hline
  \end{tabular}
\end{table}

\newpage
\begin{table}[!htb]
\caption{Several examples using a set of testing prompts.}\label{table:examples2}
\begin{tabular}{l m{2cm} c} \hline
Image                   & Prompt & Generated text \\ \hline
\multirow{3}{*}{\includegraphics[width=0.35\linewidth]{images/image1.png}} &  \footnotesize{Individual wearing a brown shirt}   & \footnotesize{
    \begin{lstlisting}[language=Python,showstringspaces=false,]   
{"Asian","F","20-39"}
    \end{lstlisting}}  \\  \cline{2-3}
                        &  \footnotesize{Individual wearing a headscarf}   & \footnotesize{
    \begin{lstlisting}[language=Python,showstringspaces=false,]   
{"M. Eastern","F","20-39"}
    \end{lstlisting}}  \\ \cline{2-3}
                        &  \footnotesize{Individual wearing eye glasses}   & \footnotesize{
    \begin{lstlisting}[language=Python,showstringspaces=false,]   
[{"Black","F","60+"},
{"White","F","40-59"}]
    \end{lstlisting}}  \\ \hline
\begin{minipage}{.35\textwidth}
    \centering\includegraphics[width=\linewidth]{images/image9.png}
    \end{minipage} &  \footnotesize{Calculate the number of Indian individuals}   & \footnotesize{Seven}  \\ \hline  
\multirow{2}{*}{\includegraphics[width=0.35\linewidth]{images/image4.png}} &  \footnotesize{Individual wearing a red jacket} & \footnotesize{
    \begin{lstlisting}[language=Python,showstringspaces=false,]   
{"M. Eastern","F","20-39"}
    \end{lstlisting}}  \\  \cline{2-3}
                        &  \footnotesize{\vspace{20pt}Male Individual}   & \vspace{10pt}\footnotesize{
    \begin{lstlisting}[language=Python,showstringspaces=false,]   
[{"Black","M","20-39"},
{"White","M","20-39"}]
    \end{lstlisting}}  \\ 
    &  &  \\ \hline
\multirow{3}{*}{\includegraphics[width=0.35\linewidth]{images/image3.png}} &  \footnotesize{How many males are in the image?}   & \footnotesize{There are three males in the image}  \\  \cline{2-3}
                        &  \footnotesize{Calculate the number of Asians}   & \footnotesize{Two}  \\ \cline{2-3}
                        &  \footnotesize{Calculate the number of Black individuals}   & \footnotesize{One}  \\ \hline

\end{tabular}
\end{table}

\newpage
\begin{table}[!htb]
    \begin{tblr}{
      colspec = {*{3}{X[l]} *{2}{X[2,l]} },
      stretch = 1,
      hlines,
    }
     \SetCell[r=2]{c} \includegraphics[scale=0.55]{images/image8.png} & \footnotesize{How many individuals are under 10 in the image} & \footnotesize{There are two individuals who appear to be under the age of 10 in the image} \\
       & \footnotesize{How many individuals are more than 60 in the image} & \footnotesize{There are two individuals who appear to be older adults and may be over the age of 60} \\ 
       \SetCell[r=2]{c} \includegraphics[scale=0.5]{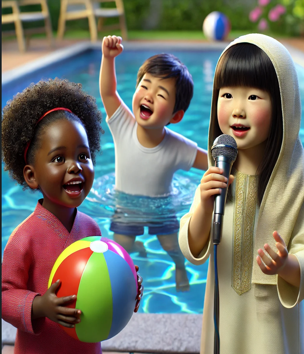} & \footnotesize{What is the race, gender, and age group of the person holding a ball?} & \footnotesize{Black, Female, 0-9}  \\
       & \footnotesize{What is the race, gender, and age group of the person singing?} & \footnotesize{Asian, Female, 0-9} \\
       \SetCell[r=2]{c} \includegraphics[scale=0.6]{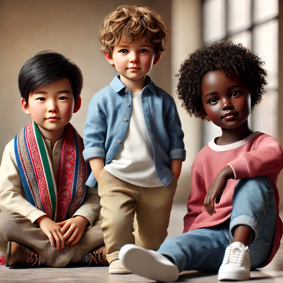} & \footnotesize{What is the race, gender, and age group of the person bending their knee?} & \footnotesize{Black, Female, 0-9}  \\
       & \footnotesize{What is the race, gender, and age group of the person standing?} & \footnotesize{White, Male, 0-9} \\
       \SetCell[r=3]{c} \includegraphics[scale=0.5]{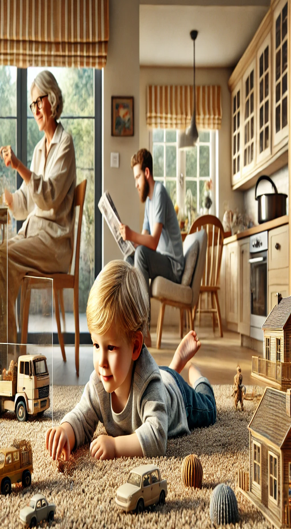} & \footnotesize{What is the race, gender, and age group of the person with the newspaper?} & \footnotesize{White, Male, 20-39}  \\
       & \footnotesize{What is the race, gender, and age group of the person on the carpet?} & \footnotesize{White, Male, 0-9} \\
       & \footnotesize{What is the race, gender, and age group of the white haired person?} & \footnotesize{White, Female, More than 60} \\
       \SetCell[r=2]{c} \includegraphics[scale=0.6]{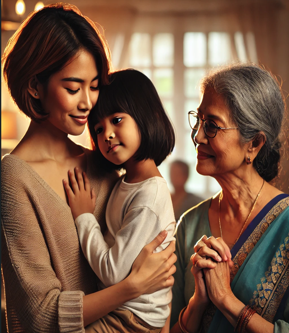} & \footnotesize{What is the race, gender, and age group of the person with the crossed fingers on both hands?} & \footnotesize{Asian, Female, More than 60}  \\
       & \footnotesize{What is the race, gender, and age group of the person holding a child?} & \footnotesize{Asian, Female, 20-39} \\
    \end{tblr}
\end{table}


\begin{table}[!htb]
    \caption{Several examples using emotion testing prompt.}\label{table:examples3}
    \begin{tblr}{
      colspec = {*{2}{X[l]} *{2}{X[2,l]} },
      stretch = 1,
      hlines,
    }
     Image & Prompt & Generated text \\
      \SetCell[r=2]{c} \includegraphics[scale=0.62]{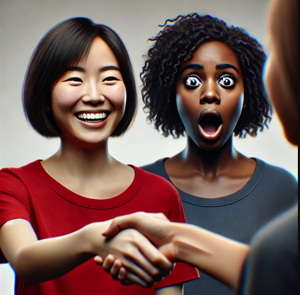} & \footnotesize{Emotion of each
person} & \footnotesize{The person on the left appears to be happy. The person in the middle appears to be surprised. The person on the right's emotion is not visible.} \\
       & \footnotesize{Emotion of the individual wearing the grey shirt} & \footnotesize{Surprise} \\ 
       \SetCell[r=2]{c} \includegraphics[scale=0.57]{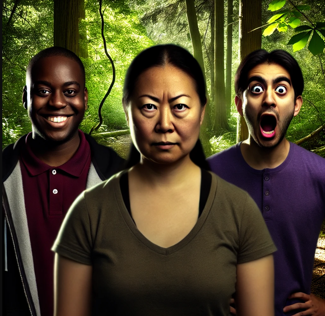} & \footnotesize{Emotion of each
person} & \footnotesize{The person on the left appears to be expressing ``happy''. The person in the center seems to be expressing ``anger''. The person on the right appears to be exhibiting ``surprise''.}  \\
       & \footnotesize{Emotion of the Black individual} & \footnotesize{Happy} \\
       \SetCell[r=2]{c} \includegraphics[scale=0.56]{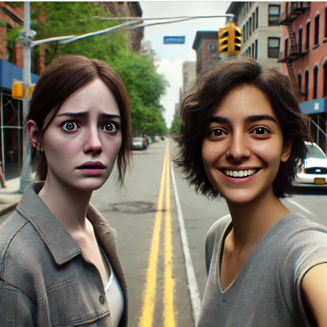} & \footnotesize{Emotion of each
person} & \footnotesize{The person on the left appears to be experiencing `fear', while the person on the right appears to be `happy'.}  \\
       & \footnotesize{Emotion of the Middle Eastern individual} & \footnotesize{Happy} \\
       \SetCell[r=2]{c} \includegraphics[scale=0.5]{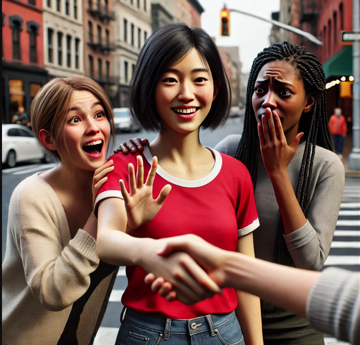} & \footnotesize{Emotion of each
person} & \footnotesize{Person on the left: surprise. Person in the middle: happy. Person on the right: fear.}  \\
       & \footnotesize{Emotion of the Asian Eastern individual} & \footnotesize{Happy} \\
       \SetCell[r=2]{c} \includegraphics[scale=0.41]{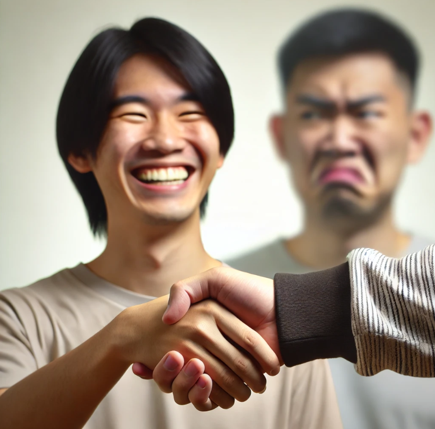} & & \\ & \footnotesize{Emotion of each
person} & \footnotesize{Person in the foreground: happy. Person in the background: disgust}  \\ 
    \end{tblr}
\end{table}

\newpage
\bibliographystyle{naturemag}
\bibliography{main}

\end{document}